\documentclass[10pt, a4paper]{article}

\usepackage{lrec2026}

\usepackage{listings}

\title{MultiWikiQA: A Reading Comprehension Benchmark in 300+ Languages}

\name{Dan Saattrup Smart} 

\address{
    Alexandra Institute \\
    Rued Langgaards Vej 7, 5., 2300 Copenhagen S, Denmark \\
    dan.smart@alexandra.dk\\
}

\abstract{
    We introduce a new reading comprehension dataset, dubbed MultiWikiQA, which covers 306 languages and has 1,220,757 samples in total.
    We start with Wikipedia articles, which also provide the context for the dataset samples, and use an LLM to generate question/answer pairs related to the Wikipedia article, ensuring that the answer appears verbatim within the article. Next, the question is then rephrased to hinder simple word matching methods from performing well on the dataset. 
    We conduct a crowdsourced human evaluation of the fluency of the generated questions, which included 156 respondents across 30 of the languages (both low- and high-resource). All 30 languages received a mean fluency rating above ``mostly natural'', showing that the samples are of good quality.
    We evaluate 6 different language models, both decoder and encoder models of varying sizes, showing that the benchmark is sufficiently difficult and that there is a large performance discrepancy amongst the languages. Both the dataset and survey evaluations are publicly available.
}

\begin{document}

\maketitleabstract

\section{Introduction}
Extracting information from documents is one of the primary uses of large language models (LLMs), especially with the rise of retrieval-augmented generation (RAG) use cases. Reading comprehension, also known as extractive question answering, is a key component in such information extraction. At its core, it concerns locating an answer to the user's query within the provided document.

This relevance of reading comprehension tasks to downstream use cases also increases the importance of having access to high-quality reading comprehension evaluation datasets within all languages.

In this work, we generate a reading comprehension dataset for 306 different languages, based on Wikipedia articles, thus increasing the access to evaluation datasets within all of these languages. The questions are all generated with an LLM, and we evaluate the quality of the generated questions within 30 of the languages through crowdsourcing. Lastly, we evaluate several language models on all of the languages, mapping out the performance of these models across a wide variety of languages. Our key contributions are:

\begin{enumerate}
    \item Release of a multilingual reading comprehension dataset in 306 languages for evaluation of  encoder, decoder and encoder-decoder language models.\footnote{The dataset can be found at \url{https://hf.co/datasets/alexandrainst/multi-wiki-qa}.}.
    \item Results and raw data from 156 crowdsourced quality evaluations of the LLM-generated questions within the dataset, across 30 languages.\footnote{The raw quality evaluation survey data can be found at \url{https://bit.ly/multi-wiki-qa-fluency-annotations}.}.
    \item Evaluations of 6 different language models on 261 languages.
\end{enumerate}

\begin{figure*}[!htb]
    \centering
    \includegraphics[width=0.9\linewidth]{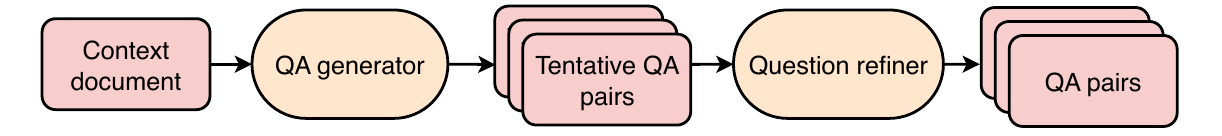}
    \caption{MultiWikiQA dataset generation process.}
    \label{fig:dataset-generation}
\end{figure*}

\section{Related Work}

Many reading comprehension datasets have been published in different languages, including English \citelanguageresource{rajpurkar-etal-2016-squad,kwiatkowski-etal-2019-natural,joshi-etal-2017-triviaqa}, Polish \citelanguageresource{rybak-etal-2024-polqa}, Korean \citelanguageresource{jun-etal-2022-korean}, Norwegian \citelanguageresource{ivanova-etal-2023-norquad,liu-etal-2024-nlebench}, German \citelanguageresource{moller-etal-2021-germanquad}, French \citelanguageresource{dhoffschmidt-etal-2020-fquad}, Icelandic \citelanguageresource{snaebjarnarson-einarsson-2022-natural}, Faroese \citelanguageresource{simonsen-etal-2025-foqa} and Russian \citelanguageresource{efimov2020sberquad}. Multilingual reading comprehension datasets have also been released, covering 7 languages \citelanguageresource{lewis-etal-2020-mlqa}, 26 languages \citelanguageresource{longpre-etal-2021-mkqa}, 11 languages \citelanguageresource{clark-etal-2020-tydi}, and 3 languages \citelanguageresource{nielsen-2023-scandeval}.

All of these benchmarks only cover a small fraction of the world's written languages, leaving most of the low-resource languages behind. Belebele \citelanguageresource{bandarkar-etal-2024-belebele}, a multiple-choice reading comprehension dataset, is a notable exception, which spans an impressive 122 languages. The multiple-choice format of Belebele is quite different compared to regular extractive question answering datasets, however. Furthermore, it is abstractive, contains only 900 samples for each language, and these samples only have short contexts of approximately 500 characters.

\section{Methodology}
\label{sec:methodology}

\subsection{Dataset Generation}
\label{sec:dataset-generation}

The dataset generation methodology closely follows the methodology in \citetlanguageresource{simonsen-etal-2025-foqa}, albeit with minor tweaks and using a different LLM, which we will describe below - it is also illustrated in Figure~\ref{fig:dataset-generation}.

From a given document, we start by generating tentative questions and answers with the LLM using the system and user prompt in Figure~\ref{fig:generation-prompt}. We ask the model to generate 2-10 different question-answer pairs for each article, both for efficiency and diversity reasons, since it is more likely to generate new question-answer pairs when conditioned on the previously generated pairs. We use structured generation to ensure a valid JSON output.

\begin{figure}
\scriptsize
\begin{verbatim}
SYSTEM:
You are a helpful {language} question answering 
dataset generator. The only language you know is 
{language}.

USER:
The following is a Wikipedia article in {language}.

<article>
{article}
</article>

Generate 2 to 10 questions about the article, 
depending on the length of the article, all 
of which answered in the article.

You also have to supply answers to the questions, 
and the answers have to appear exactly as written
in the article (including same casing).

The answers should only contain the answers 
themselves, and not the surrounding sentence - 
keep the answers as short as possible.

The answers have to be different from each other.

All your questions and answers must be in 
{language}.

Your answer must be a JSON dictionary with the 
key "results", with the value being a list of 
dictionaries having keys "question" and "answer".
\end{verbatim}
\caption{The system and user prompt used to generate the tentative questions and answers.}
\label{fig:generation-prompt}
\end{figure}

Next, we filter the generated JSON dictionaries by checking that each question-answer entry contains the appropriate ``question'' and ``answer'' keys, as well as checking if the answer appears verbatim in the context document.

We \textit{could} stop at this point, as we now have a set of questions and answers for the context document. However, other reading comprehension datasets have been criticised for having questions that used the same wording as the context document, making it too easy for language models to ``cheat'' by simply word matching \cite{weissenborn2017making, jia2017adversarial}. In an attempt to prevent this, we proceed with a separate rephrasing stage, where we prompt the same LLM to rephrase the question (without the context), using the prompt in Figure~\ref{fig:rephrasing-prompt}.

\begin{figure}
\scriptsize
\begin{verbatim}
The following is a {language} question.

<question>
{question}
</question>

Re-write the question as much as possible, 
preserving the meaning, using synonyms, 
other phrases, or a different (valid) 
word order.

Your question must be in {language}.

Your answer must be a JSON dictionary with the 
key "question".
\end{verbatim}
\caption{The prompt used to rephrase the generated tentative questions.}
\label{fig:rephrasing-prompt}
\end{figure}

The resulting set of context-question-answer triples are then collected into a dataset of the same format as SQuAD~\citelanguageresource{rajpurkar-etal-2016-squad}.

\subsection{Quality Evaluation of LLM-generated Questions}
\label{sec:quality-evaluation}

To evaluate the quality of the LLM-generated questions, we conducted a survey in all the included languages, and crowdsourced answers from various social media channels. Each survey contained a random sample of 50 generated questions for the given language, and prompted the user to rate the fluency of each question as 1, 2, or 3 stars. The precise preamble that was presented to each user can be seen in Figure~\ref{fig:survey-preamble}.

\begin{figure*}[!htb]
    \begin{itshape}
        \small
        Please indicate how natural the following questions are:
    
        \begin{itemize}
            \item[$\star$] Does not sound natural at all
            \item[$\star\star$] Sounds mostly natural, but there is a particular part of the question that looks wrong
            \item[$\star\star\star$] Sounds like a natural question
        \end{itemize}
    
        Note that ``naturalness'' here is only meaning fluency, so whether the question is unanswerable or requires context to be answered does not matter here.
    \end{itshape}
    \caption{The preamble used in all of the surveys.}
    \label{fig:survey-preamble}
\end{figure*}

We used the Microsoft Forms service\footnote{\url{https://forms.cloud.microsoft/}} to facilitate the individual language surveys, and we self-hosted a simple routing interface which guided users to the correct language survey - the interface can be seen in the appendix. As each survey was created manually, we did not create the surveys for all languages, but instead had the routing interface send us an email if the user selected a language not currently covered - we then sent an email reply to the user when the language was included. The routing interface was coded using Vue.js \cite{you2025vuejs} - the source code can be found in the appendix.

\begin{figure*}[!htb]
    \centering
    \includegraphics[width=1\linewidth]{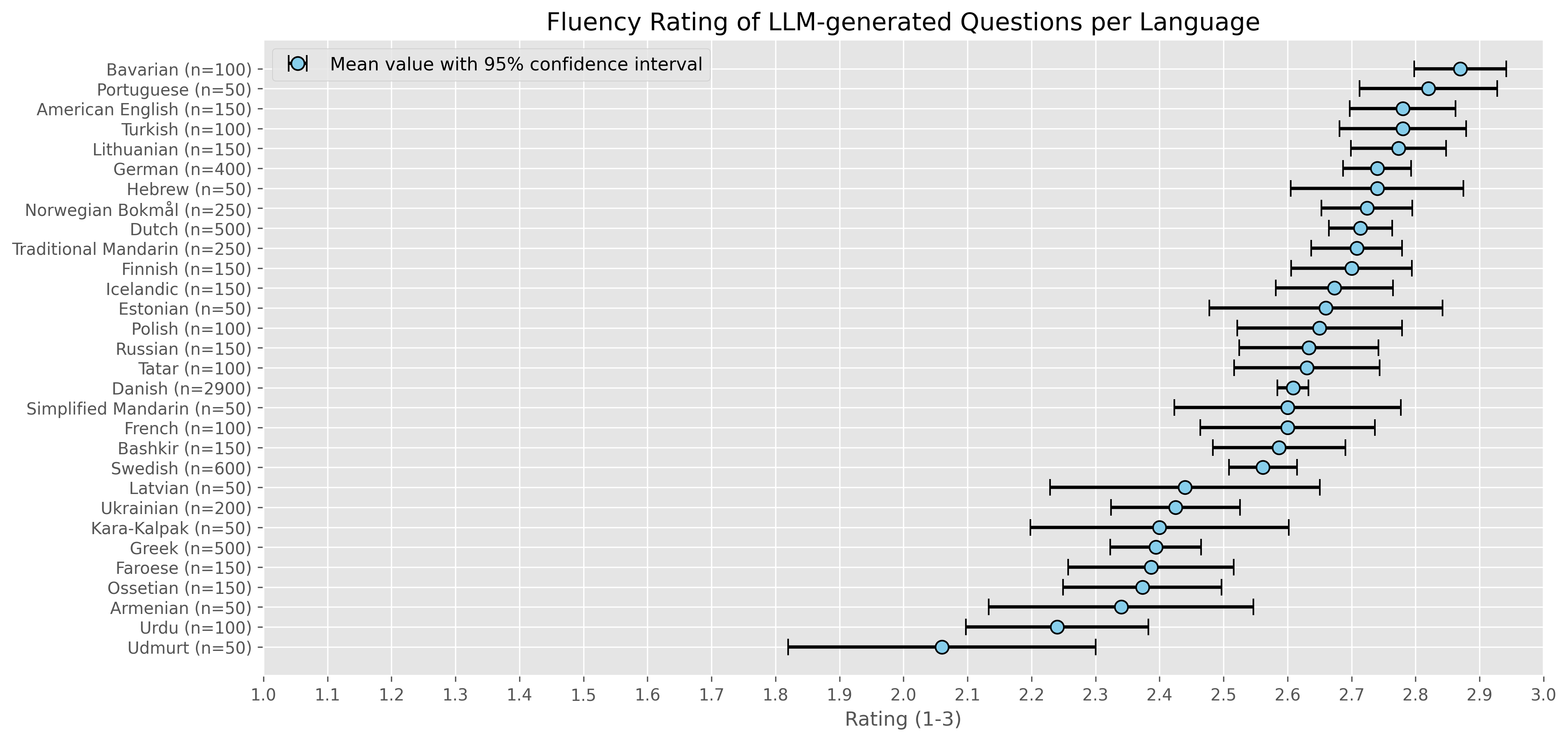}
    \caption{Results from the conducted fluency surveys.}
    \label{fig:quality-evaluation-scores}
\end{figure*}

\section{The Dataset}
\label{sec:the-dataset}

We generate the dataset using the methodology in Section~\ref{sec:dataset-generation}. We use the 20231101 Wikipedia dump\footnote{\url{https://hf.co/datasets/wikimedia/wikipedia}} and include 306 of the Wikipedia languages - a full list can be found in the appendix. We include special cases for Mandarin and Portuguese. We split Mandarin articles into Simplified Mandarin (``zh-cn'') and Traditional Mandarin (``zh-tw'') using the HanzIdentifier~\cite{tsroten2024hanzidentifier}, and we split Portuguese articles into European Portuguese (``pt-pt'') and Brazilian Portuguese (``pt-br'') using the PtVid classifier~\citelanguageresource{sousa2025enhancing}.

We use the Gemini-1.5-pro model \cite{reid2024gemini} for the question generation with temperature 1.0 and where we allow 1,000 generated tokens, and we stop generating for a given language when we reach 5,000 context-question-answer samples, or when we run out of articles. This yielded 1,220,757 samples in total. We ran out of articles for 101 languages - see the appendix for an overview of these languages.

Using the question evaluation methodology in Section~\ref{sec:quality-evaluation}, we get 156 survey respondents in 30 different languages. The mean quality scores across the languages, along with the number of survey responses, can be found in Figure~\ref{fig:quality-evaluation-scores}. We see that the generated questions have a mean rating above 2.0, corresponding to ``mostly natural'', even for the languages Bashkir, Kara-Kalpak, Faroese, Ossetian, Udmurt and Icelandic, all having fewer than one million native speakers.

\section{Evaluations}
\label{sec:evaluations}

We evaluate a variety of language models on the MultiWikiQA dataset in all the languages with at least 1,024 samples for training, 32 for validation and 128 for testing. This was chosen as we are also evaluating encoder models, and 1,024 samples was found in \citet{nielsen-2023-scandeval} to be enough for the models to adequately fit the data for several reading comprehension datasets and languages. There were 261 languages satisfying this criterion.

The evaluation itself was conducted using the EuroEval framework \cite{nielsen-2023-scandeval,saattrup-nielsen-etal-2025-encoder}. See the list of evaluated models in Table~\ref{tab:evaluated-models}. The decoder models were evaluated 2-shot, which was preferred over zero-shot evaluation to enable proper evaluation of base decoder models. The few-shot examples come from the training split. The encoder models were trained on the training split, with early stopping based on the validation split, and the final performance reported on the test split.

The results are visualised in Figure~\ref{fig:evaluation-results}, and the full results can be found in the appendix. From the results, we see that there is a large discrepancy in performance across languages, which is consistent across the three different model types.

\begin{table*}[!htb]
    \centering
    \small
    \begin{tabular}{lclr}
        \hline
        \textbf{Model Name} & \textbf{Parameters} & \textbf{Type} & \textbf{Mean F1-score} \\
        \hline
        Mistral-Small-3.1-24B-Instruct-2503 \cite{mistralsmall2025} & 24B & Instruct Decoder & 55.83\% ± 1.09\% \\
        Mistral-Small-3.1-24B-Base-2503 \cite{mistralsmall2025} & 24B & Base Decoder & 54.71\% ± 1.20\% \\
        Llama-3.1-8B-Instruct \cite{grattafiori2024llama} & 8B & Instruct Decoder & 52.38\% ± 0.91\% \\
        Llama-3.1-8B \cite{grattafiori2024llama} & 8B & Base Decoder & 47.26\% ± 1.22\% \\
        Multilingual-E5-large \cite{wang2024multilingual} & 560M & Encoder & 23.82\% ± 0.65\% \\
        XLM-RoBERTa-large \cite{ruder2019unsupervised} & 561M & Encoder & 20.23\% ± 0.69\% \\
        \hline
    \end{tabular}
    \caption{Evaluation results on MultiWikiQA in 261 languages. See the raw results in the appendix.}
    \label{tab:evaluated-models}
\end{table*}

\begin{figure*}[!htb]
    \centering
    \includegraphics[width=\textwidth]{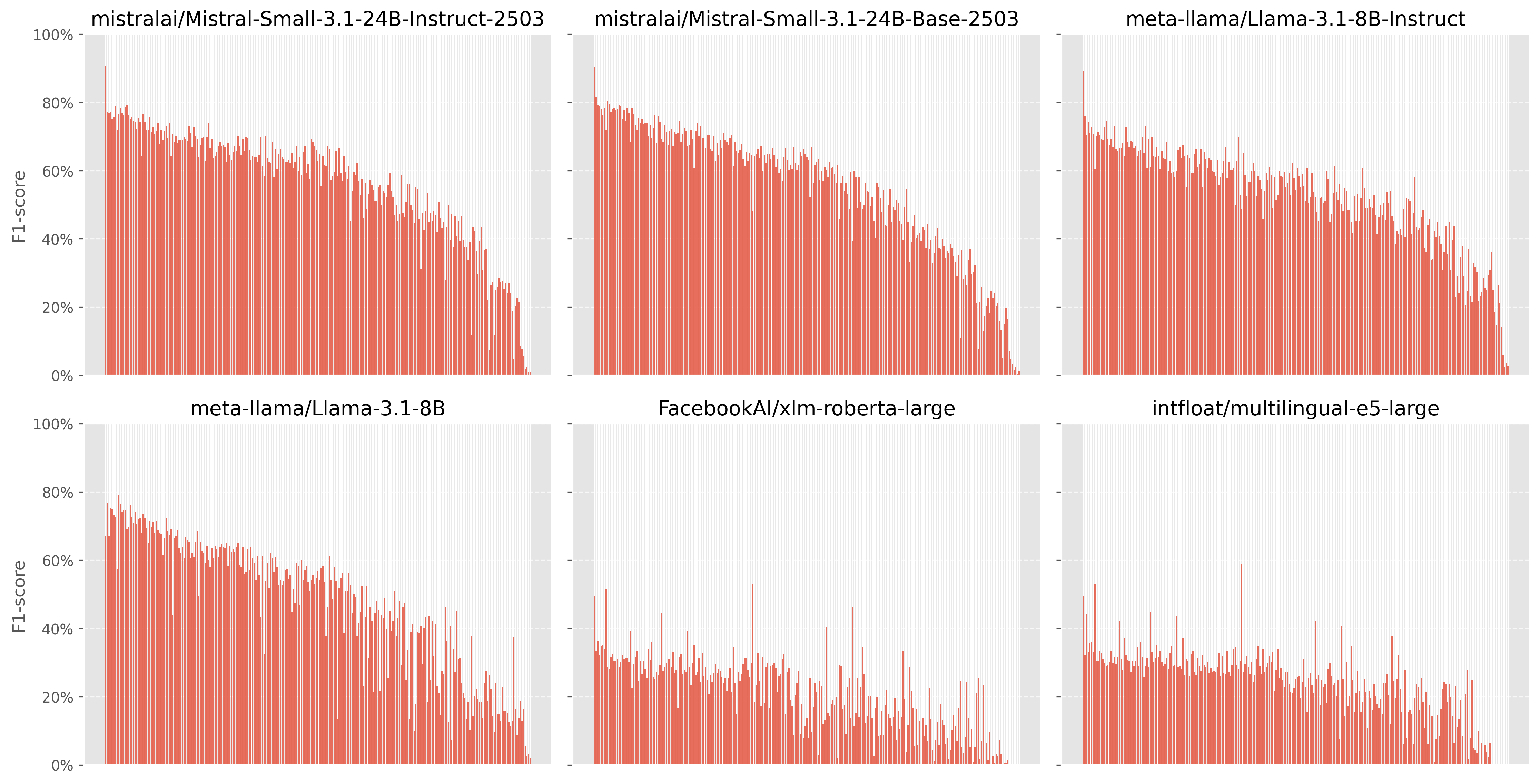}
    \caption{F1-score performance of six models across 261 languages, where the languages on the x-axis are sorted in descending order based on the mean F1-score across all models. See the raw results in the appendix.}
    \label{fig:evaluation-results}
\end{figure*}

\section{Conclusion}
\label{sec:conclusion}

We have introduced a new reading comprehension dataset in 306 languages, based on Wikipedia articles, with questions generated by an LLM. We crowdsourced the quality of the LLM-generated questions for 30 languages, showing that the LLM-generated questions are of good quality. Lastly, we evaluated 6 models on all of the languages, which showed that the benchmark is sufficiently difficult across instruction-tuned decoders, base decoders and encoders, and that there is a large performance discrepancy amongst the languages.

\section*{Limitations}
\label{sec:limitations}

While we got survey responses in 30 different languages, that still only covers approximately 10\% of the languages covered in the dataset, so we cannot guarantee that the conclusions from the surveys generalise to the remaining languages. However, since the surveys cover a wide spectrum of language families and language resource levels, we are quite confident in such a generalisation.

\section*{Acknowledgements}
\label{sec:acknowledgements}

This research was funded by the EU Horizon project TrustLLM (grant agreement number 101135671) and the LLM credits used to generate the dataset with the Gemini-1.5-pro model were funded as part of the Google Cloud Research Credits Programme.

\newpage
\nocite{*}
\section{Bibliographical References}\label{sec:reference}

\bibliographystyle{lrec2026-natbib}
\bibliography{bib}

\begin{thebibliography}{18}
\expandafter\ifx\csname natexlab\endcsname\relax\def\natexlab#1{#1}\fi

\bibitem[{Bandarkar et~al.(2024)Bandarkar, Liang, Muller, Artetxe, Shukla, Husa, Goyal, Krishnan, Zettlemoyer, and Khabsa}]{bandarkar-etal-2024-belebele}
Bandarkar, Lucas and Liang, Davis and Muller, Benjamin and Artetxe, Mikel and Shukla, Satya Narayan and Husa, Donald and Goyal, Naman and Krishnan, Abhinandan and Zettlemoyer, Luke and Khabsa, Madian. 2024.
\newblock \href {https://doi.org/10.18653/v1/2024.acl-long.44} {\emph{The Belebele Benchmark: a Parallel Reading Comprehension Dataset in 122 Language Variants}}.
\newblock Association for Computational Linguistics.
\newblock PID \href{https://hf.co/datasets/facebook/belebele}{https://hf.co/datasets/facebook/belebele}.

\bibitem[{Clark et~al.(2020)Clark, Choi, Collins, Garrette, Kwiatkowski, Nikolaev, and Palomaki}]{clark-etal-2020-tydi}
Clark, Jonathan H. and Choi, Eunsol and Collins, Michael and Garrette, Dan and Kwiatkowski, Tom and Nikolaev, Vitaly and Palomaki, Jennimaria. 2020.
\newblock \href {https://doi.org/10.1162/tacl_a_00317} {\emph{{T}y{D}i {QA}: A Benchmark for Information-Seeking Question Answering in Typologically Diverse Languages}}.
\newblock MIT Press.
\newblock PID \href{https://hf.co/datasets/google-research-datasets/tydiqa}{https://hf.co/datasets/google-research-datasets/tydiqa}.

\bibitem[{d{'}Hoffschmidt et~al.(2020)d{'}Hoffschmidt, Belblidia, Heinrich, Brendl{\'e}, and Vidal}]{dhoffschmidt-etal-2020-fquad}
d{'}Hoffschmidt, Martin and Belblidia, Wacim and Heinrich, Quentin and Brendl{\'e}, Tom and Vidal, Maxime. 2020.
\newblock \href {https://doi.org/10.18653/v1/2020.findings-emnlp.107} {\emph{{FQ}u{AD}: {F}rench Question Answering Dataset}}.
\newblock Association for Computational Linguistics.
\newblock PID \href{https://hf.co/datasets/illuin/fquad}{https://hf.co/datasets/illuin/fquad}.

\bibitem[{Efimov et~al.(2020)Efimov, Chertok, Boytsov, and Braslavski}]{efimov2020sberquad}
Efimov, Pavel and Chertok, Andrey and Boytsov, Leonid and Braslavski, Pavel. 2020.
\newblock \emph{SberQuAD--Russian Reading Comprehension Dataset: Description and Analysis}.
\newblock Springer.
\newblock PID \href{http://hf.co/datasets/kuznetsoffandrey/sberquad}{http://hf.co/datasets/kuznetsoffandrey/sberquad}.

\bibitem[{Ivanova et~al.(2023)Ivanova, Andreassen, Jentoft, Wold, and {\O}vrelid}]{ivanova-etal-2023-norquad}
Ivanova, Sardana and Andreassen, Fredrik and Jentoft, Matias and Wold, Sondre and {\O}vrelid, Lilja. 2023.
\newblock \href {https://aclanthology.org/2023.nodalida-1.17/} {\emph{{N}or{Q}u{AD}: {N}orwegian Question Answering Dataset}}.
\newblock University of Tartu Library.
\newblock PID \href{https://hf.co/datasets/ltg/norquad}{https://hf.co/datasets/ltg/norquad}.

\bibitem[{Joshi et~al.(2017)Joshi, Choi, Weld, and Zettlemoyer}]{joshi-etal-2017-triviaqa}
Joshi, Mandar and Choi, Eunsol and Weld, Daniel and Zettlemoyer, Luke. 2017.
\newblock \href {https://doi.org/10.18653/v1/P17-1147} {\emph{{T}rivia{QA}: A Large Scale Distantly Supervised Challenge Dataset for Reading Comprehension}}.
\newblock Association for Computational Linguistics.
\newblock PID \href{https://hf.co/datasets/mandarjoshi/trivia_qa}{https://hf.co/datasets/mandarjoshi/trivia\_qa}.

\bibitem[{Jun et~al.(2022)Jun, Choi, Sim, Kim, Jang, and Min}]{jun-etal-2022-korean}
Jun, Changwook and Choi, Jooyoung and Sim, Myoseop and Kim, Hyun and Jang, Hansol and Min, Kyungkoo. 2022.
\newblock \href {https://aclanthology.org/2022.lrec-1.657/} {\emph{{K}orean-Specific Dataset for Table Question Answering}}.
\newblock European Language Resources Association.
\newblock PID \href{https://hf.co/datasets/KorQuAD/squad_kor_v1}{https://hf.co/datasets/KorQuAD/squad\_kor\_v1}.

\bibitem[{Kwiatkowski et~al.(2019)Kwiatkowski, Palomaki, Redfield, Collins, Parikh, Alberti, Epstein, Polosukhin, Devlin, Lee, Toutanova, Jones, Kelcey, Chang, Dai, Uszkoreit, Le, and Petrov}]{kwiatkowski-etal-2019-natural}
Kwiatkowski, Tom and Palomaki, Jennimaria and Redfield, Olivia and Collins, Michael and Parikh, Ankur and Alberti, Chris and Epstein, Danielle and Polosukhin, Illia and Devlin, Jacob and Lee, Kenton and Toutanova, Kristina and Jones, Llion and Kelcey, Matthew and Chang, Ming-Wei and Dai, Andrew M. and Uszkoreit, Jakob and Le, Quoc and Petrov, Slav. 2019.
\newblock \href {https://doi.org/10.1162/tacl_a_00276} {\emph{Natural Questions: A Benchmark for Question Answering Research}}.
\newblock MIT Press.
\newblock PID \href{https://hf.co/datasets/google-research-datasets/natural_questions}{https://hf.co/datasets/google-research-datasets/natural\_questions}.

\bibitem[{Lewis et~al.(2020)Lewis, Oguz, Rinott, Riedel, and Schwenk}]{lewis-etal-2020-mlqa}
Lewis, Patrick and Oguz, Barlas and Rinott, Ruty and Riedel, Sebastian and Schwenk, Holger. 2020.
\newblock \href {https://doi.org/10.18653/v1/2020.acl-main.653} {\emph{{MLQA}: Evaluating Cross-lingual Extractive Question Answering}}.
\newblock Association for Computational Linguistics.
\newblock PID \href{https://hf.co/datasets/facebook/mlqa}{https://hf.co/datasets/facebook/mlqa}.

\bibitem[{Liu et~al.(2024)Liu, Zhang, Farup, Lauvrak, Ingvaldsen, Eide, Gulla, and Yang}]{liu-etal-2024-nlebench}
Liu, Peng and Zhang, Lemei and Farup, Terje and Lauvrak, Even W. and Ingvaldsen, Jon Espen and Eide, Simen and Gulla, Jon Atle and Yang, Zhirong. 2024.
\newblock \href {https://doi.org/10.18653/v1/2024.emnlp-main.317} {\emph{{NLEB}ench+{N}or{GLM}: A Comprehensive Empirical Analysis and Benchmark Dataset for Generative Language Models in {N}orwegian}}.
\newblock Association for Computational Linguistics.
\newblock PID \href{https://hf.co/datasets/NorGLM/NO-Multi-QA-Sum}{https://hf.co/datasets/NorGLM/NO-Multi-QA-Sum}.

\bibitem[{Longpre et~al.(2021)Longpre, Lu, and Daiber}]{longpre-etal-2021-mkqa}
Longpre, Shayne and Lu, Yi and Daiber, Joachim. 2021.
\newblock \href {https://doi.org/10.1162/tacl_a_00433} {\emph{{MKQA}: A Linguistically Diverse Benchmark for Multilingual Open Domain Question Answering}}.
\newblock MIT Press.
\newblock PID \href{https://hf.co/datasets/apple/mkqa}{https://hf.co/datasets/apple/mkqa}.

\bibitem[{M{\"o}ller et~al.(2021)M{\"o}ller, Risch, and Pietsch}]{moller-etal-2021-germanquad}
M{\"o}ller, Timo and Risch, Julian and Pietsch, Malte. 2021.
\newblock \href {https://doi.org/10.18653/v1/2021.mrqa-1.4} {\emph{{G}erman{Q}u{AD} and {G}erman{DPR}: Improving Non-{E}nglish Question Answering and Passage Retrieval}}.
\newblock Association for Computational Linguistics.
\newblock PID \href{https://hf.co/datasets/deepset/germanquad}{https://hf.co/datasets/deepset/germanquad}.

\bibitem[{Nielsen(2023)}]{nielsen-2023-scandeval}
Nielsen, Dan. 2023.
\newblock \href {https://aclanthology.org/2023.nodalida-1.20/} {\emph{{S}cand{E}val: A Benchmark for {S}candinavian Natural Language Processing}}.
\newblock University of Tartu Library.
\newblock PID \href{https://doi.org/10.57967/hf/6061}{https://doi.org/10.57967/hf/6061}.

\bibitem[{Rajpurkar et~al.(2016)Rajpurkar, Zhang, Lopyrev, and Liang}]{rajpurkar-etal-2016-squad}
Rajpurkar, Pranav and Zhang, Jian and Lopyrev, Konstantin and Liang, Percy. 2016.
\newblock \href {https://doi.org/10.18653/v1/D16-1264} {\emph{{SQ}u{AD}: 100,000+ Questions for Machine Comprehension of Text}}.
\newblock Association for Computational Linguistics.
\newblock PID \href{https://hf.co/datasets/rajpurkar/squad}{https://hf.co/datasets/rajpurkar/squad}.

\bibitem[{Rybak et~al.(2024)Rybak, Przyby{\l}a, and Ogrodniczuk}]{rybak-etal-2024-polqa}
Rybak, Piotr and Przyby{\l}a, Piotr and Ogrodniczuk, Maciej. 2024.
\newblock \href {https://aclanthology.org/2024.lrec-main.1125/} {\emph{{P}ol{QA}: {P}olish Question Answering Dataset}}.
\newblock ELRA and ICCL.
\newblock PID \href{https://hf.co/datasets/ipipan/polqa}{https://hf.co/datasets/ipipan/polqa}.

\bibitem[{Simonsen et~al.(2025)Simonsen, Nielsen, and Einarsson}]{simonsen-etal-2025-foqa}
Simonsen, Annika and Nielsen, Dan Saattrup and Einarsson, Hafsteinn. 2025.
\newblock \href {https://aclanthology.org/2025.resourceful-1.11/} {\emph{{F}o{QA}: A {F}aroese Question-Answering Dataset}}.
\newblock University of Tartu Library, Estonia.
\newblock PID \href{https://doi.org/10.57967/hf/5615}{https://doi.org/10.57967/hf/5615}.

\bibitem[{Sn{\ae}bjarnarson and Einarsson(2022)}]{snaebjarnarson-einarsson-2022-natural}
Sn{\ae}bjarnarson, V{\'e}steinn and Einarsson, Hafsteinn. 2022.
\newblock \href {https://aclanthology.org/2022.lrec-1.477/} {\emph{Natural Questions in {I}celandic}}.
\newblock European Language Resources Association.
\newblock PID \href{https://hf.co/datasets/vesteinn/icelandic-qa-NQiI}{https://hf.co/datasets/vesteinn/icelandic-qa-NQiI}.

\bibitem[{Sousa et~al.(2025)Sousa, Almeida, Silvano, Cantante, Campos, and Jorge}]{sousa2025enhancing}
Hugo Sousa and Rúben Almeida and Purificação Silvano and Inês Cantante and Ricardo Campos and Alipio Jorge. 2025.
\newblock \href {https://doi.org/10.1609/aaai.v39i24.34705} {\emph{Enhancing Portuguese Variety Identification with Cross-Domain Approaches}}.
\newblock AAAI.
\newblock PID \href{https://hf.co/datasets/liaad/PtBrVId}{https://hf.co/datasets/liaad/PtBrVId}.

\end{thebibliography}


\begin{thebibliography}{12}
\expandafter\ifx\csname natexlab\endcsname\relax\def\natexlab#1{#1}\fi

\bibitem[{Comanici et~al.(2025)Comanici, Bieber, Schaekermann, Pasupat, Sachdeva, Dhillon, Blistein, Ram, Zhang, Rosen et~al.}]{comanici2025gemini}
Gheorghe Comanici, Eric Bieber, Mike Schaekermann, Ice Pasupat, Noveen Sachdeva, Inderjit Dhillon, Marcel Blistein, Ori Ram, Dan Zhang, Evan Rosen, et~al. 2025.
\newblock Gemini 2.5: Pushing the frontier with advanced reasoning, multimodality, long context, and next generation agentic capabilities.
\newblock \emph{arXiv preprint arXiv:2507.06261}.

\bibitem[{Grattafiori et~al.(2024)Grattafiori, Dubey, Jauhri, Pandey, Kadian, Al-Dahle, Letman, Mathur, Schelten, Vaughan et~al.}]{grattafiori2024llama}
Aaron Grattafiori, Abhimanyu Dubey, Abhinav Jauhri, Abhinav Pandey, Abhishek Kadian, Ahmad Al-Dahle, Aiesha Letman, Akhil Mathur, Alan Schelten, Alex Vaughan, et~al. 2024.
\newblock The llama 3 herd of models.
\newblock \emph{arXiv preprint arXiv:2407.21783}.

\bibitem[{Jia and Liang(2017)}]{jia2017adversarial}
Robin Jia and Percy Liang. 2017.
\newblock Adversarial examples for evaluating reading comprehension systems.
\newblock In \emph{Proceedings of the 2017 Conference on Empirical Methods in Natural Language Processing}, pages 2021--2031.

\bibitem[{Mistral-AI(2025)}]{mistralsmall2025}
Mistral-AI. 2025.
\newblock \href {https://mistral.ai/news/mistral-small-3-1} {Mistral small 3.1}.
\newblock [Online; accessed 31. Jul. 2025].

\bibitem[{Nielsen(2023)}]{nielsen-2023-scandeval}
Dan Nielsen. 2023.
\newblock \href {https://aclanthology.org/2023.nodalida-1.20/} {{S}cand{E}val: A benchmark for {S}candinavian natural language processing}.
\newblock In \emph{Proceedings of the 24th Nordic Conference on Computational Linguistics (NoDaLiDa)}, pages 185--201, T{\'o}rshavn, Faroe Islands. University of Tartu Library.

\bibitem[{Reid et~al.(2024)Reid, Savinov, Teplyashin, Lepikhin, Lillicrap, Alayrac, Soricut, Lazaridou, Firat, Schrittwieser et~al.}]{reid2024gemini}
Machel Reid, Nikolay Savinov, Denis Teplyashin, Dmitry Lepikhin, Timothy~P Lillicrap, Jean-Baptiste Alayrac, Radu Soricut, Angeliki Lazaridou, Orhan Firat, Julian Schrittwieser, et~al. 2024.
\newblock Gemini 1.5: Unlocking multimodal understanding across millions of tokens of context.
\newblock \emph{CoRR}.

\bibitem[{Ruder et~al.(2019)Ruder, S{\o}gaard, and Vuli{\'c}}]{ruder2019unsupervised}
Sebastian Ruder, Anders S{\o}gaard, and Ivan Vuli{\'c}. 2019.
\newblock Unsupervised cross-lingual representation learning.
\newblock In \emph{Proceedings of the 57th Annual Meeting of the Association for Computational Linguistics: Tutorial Abstracts}, pages 31--38.

\bibitem[{Saattrup~Nielsen et~al.(2025)Saattrup~Nielsen, Enevoldsen, and Schneider-Kamp}]{saattrup-nielsen-etal-2025-encoder}
Dan Saattrup~Nielsen, Kenneth Enevoldsen, and Peter Schneider-Kamp. 2025.
\newblock \href {https://aclanthology.org/2025.nodalida-1.60/} {Encoder vs decoder: {Comparative} analysis of encoder and decoder language models on multilingual {NLU} tasks}.
\newblock In \emph{Proceedings of the Joint 25th Nordic Conference on Computational Linguistics and 11th Baltic Conference on Human Language Technologies (NoDaLiDa/Baltic-HLT 2025)}, pages 561--572, Tallinn, Estonia. University of Tartu Library.

\bibitem[{tsroten(2024)}]{tsroten2024hanzidentifier}
tsroten. 2024.
\newblock {hanzidentifier}.
\newblock \url{https://github.com/tsroten/hanzidentifier}.
\newblock GitHub Repository, version {v1.3.0}, accessed 2025.

\bibitem[{Wang et~al.(2024)Wang, Yang, Huang, Yang, Majumder, and Wei}]{wang2024multilingual}
Liang Wang, Nan Yang, Xiaolong Huang, Linjun Yang, Rangan Majumder, and Furu Wei. 2024.
\newblock Multilingual e5 text embeddings: A technical report.
\newblock \emph{arXiv preprint arXiv:2402.05672}.

\bibitem[{Weissenborn et~al.(2017)Weissenborn, Wiese, and Seiffe}]{weissenborn2017making}
Dirk Weissenborn, Georg Wiese, and Laura Seiffe. 2017.
\newblock Making neural qa as simple as possible but not simpler.
\newblock \emph{CoNLL 2017}, page 271.

\bibitem[{You(2025)}]{you2025vuejs}
Evan You. 2025.
\newblock {VueJS}.
\newblock \url{https://github.com/vuejs/core}.
\newblock GitHub Repository, version {v3.6.0-alpha.2}, accessed 2025.

\end{thebibliography}

\section{Language Resource References}
\label{lr:ref}
\bibliographystylelanguageresource{lrec2026-natbib}
\bibliographylanguageresource{languageresource}

\newpage
\appendix
\onecolumn

\section{Appendix}

\begin{table*}[h]
    \centering
    \scriptsize
    \begin{tabular}{ccccccccccccc}
        ab & ban & cr & fat & gur & iu & lad & mni & om & ru & ss & tt & yue \\
        ace & bar & crh & ff & guw & ja & lb & mnw & or & rue & st & tum & za \\
        ady & bcl & cs & fi & gv & jam & lbe & mr & os & rw & stq & tw & zea \\
        af & be & csb & fj & ha & jbo & lez & mrj & pa & sa & su & ty & zh-cn \\
        als & bg & cu & fo & hak & ka & lfn & ms & pag & sah & sv & tyv & zh-tw \\
        alt & bi & cv & fon & haw & kaa & lg & mt & pam & sat & sw & udm & zu \\
        am & bjn & cy & fr & he & kab & li & mwl & pap & sc & szl & ug & \\
        ami & blk & da & frp & hi & kbd & lij & my & pcd & scn & szy & uk & \\
        an & bm & dag & frr & hif & kbp & lld & myv & pcm & sco & ta & ur & \\
        ang & bn & de & fur & hr & kcg & lmo & mzn & pdc & sd & tay & uz & \\
        anp & bo & din & fy & hsb & kg & ln & nap & pfl & se & tcy & ve & \\
        ar & bpy & diq & ga & ht & ki & lo & nds & pi & sg & te & vec & \\
        arc & br & dsb & gag & hu & kk & lt & ne & pl & shi & tet & vep & \\
        ary & bs & dty & gan & hy & kl & ltg & new & pms & shn & tg & vi & \\
        arz & bug & dv & gcr & hyw & km & lv & nia & pnb & si & th & vls & \\
        as & bxr & dz & gd & ia & kn & mad & nl & pnt & sk & ti & vo & \\
        ast & ca & ee & gl & id & ko & mai & nn & ps & skr & tk & wa & \\
        atj & cdo & el & glk & ie & koi & mdf & no & pt-br & sl & tl & war & \\
        av & ce & en & gn & ig & krc & mg & nov & pt-pt & sm & tly & wo & \\
        avk & ceb & eo & gom & ik & ks & mhr & nqo & pwn & smn & tn & wuu & \\
        awa & ch & es & gor & ilo & ku & mi & nso & qu & sn & to & xal & \\
        ay & chr & et & got & inh & kv & min & nv & rm & so & tpi & xh & \\
        az & chy & eu & gpe & io & kw & mk & ny & rmy & sq & tr & xmf & \\
        azb & ckb & ext & gu & is & ky & ml & oc & rn & sr & trv & yi & \\
        ba & co & fa & guc & it & la & mn & olo & ro & srn & ts & yo & \\
    \end{tabular}
    \caption{List of all languages in MultiWikiQA.}
    \label{tab:language-overview}
\end{table*}

\begin{table*}[h]
    \centering
    \scriptsize
    \begin{tabular}{cr|cr|cr|cr}
        \hline
        \textbf{Language} & \textbf{Samples} & \textbf{Language} & \textbf{Samples} & \textbf{Language} & \textbf{Samples} & \textbf{Language} & \textbf{Samples} \\
        \hline
        lld & 4,745 & tay & 3,049 & guc & 1,558 & cu & 443 \\
        tn & 4,744 & ami & 2,920 & pwn & 1,471 & za & 427 \\
        pcm & 4,623 & nia & 2,804 & mi & 1,419 & ki & 416 \\
        gcr & 4,590 & ny & 2,751 & awa & 1,385 & tpi & 397 \\
        fat & 4,539 & ff & 2,701 & pdc & 1,381 & ti & 385 \\
        om & 4,458 & dz & 2,679 & jam & 1,345 & got & 383 \\
        av & 4,375 & shi & 2,677 & sm & 1,278 & ady & 380 \\
        se & 4,257 & wo & 2,624 & st & 1,270 & lbe & 378 \\
        tum & 4,252 & kbd & 2,585 & ee & 1,223 & ve & 369 \\
        gpe & 4,242 & bpy & 2,561 & kcg & 1,202 & srn & 321 \\
        csb & 4,199 & ln & 2,297 & to & 1,187 & kg & 261 \\
        mrj & 4,194 & ace & 2,210 & tly & 1,108 & arc & 251 \\
        gor & 4,137 & mni & 2,179 & nov & 1,066 & chr & 185 \\
        crh & 4,053 & mad & 2,032 & atj & 1,064 & bi & 149 \\
        gur & 4,042 & jbo & 2,011 & fon & 899 & iu & 148 \\
        dty & 4,029 & hak & 1,945 & nso & 810 & ch & 135 \\
        mdf & 3,957 & haw & 1,903 & pag & 802 & ty & 129 \\
        xh & 3,955 & ss & 1,856 & rn & 763 & bug & 119 \\
        krc & 3,894 & cdo & 1,780 & fj & 745 & sg & 83 \\
        frp & 3,849 & inh & 1,738 & rmy & 722 & pi & 79 \\
        guw & 3,782 & din & 1,716 & gan & 697 & ik & 67 \\
        anp & 3,667 & ltg & 1,636 & nv & 681 & cr & 33 \\
        koi & 3,539 & ab & 1,625 & bm & 663 & chy & 25 \\
        gag & 3,409 & tet & 1,604 & kl & 529 &&\\
        glk & 3,384 & ts & 1,583 & pnt & 529 &&\\
        ang & 3,376 & ks & 1,571 & xal & 464 &&
    \end{tabular}
    \caption{All the languages in MultiWikiQA with fewer than 5,000 samples.}
    \label{tab:num-samples-in-small-subsets}
\end{table*}

\begin{table*}[h]
\centering
\scriptsize
\begin{tabular}{crrrrrr}
\hline
\textbf{Language} & \textbf{Mistral Base} & \textbf{Mistral Instruct} & \textbf{Llama Base} & \textbf{Llama Instruct} & \textbf{XLM-RoBERTa} & \textbf{Multi-E5} \\
\hline
ace & 45.2\% ± 2.2\% & 49.9\% ± 4.1\% & 39.1\% ± 8.2\% & 50.5\% ± 2.5\% & 10.5\% ± 5.6\% & 20.6\% ± 13.9\% \\
af & 78.0\% ± 4.0\% & 76.5\% ± 3.1\% & 69.7\% ± 5.8\% & 74.5\% ± 1.1\% & 30.9\% ± 2.8\% & 29.1\% ± 3.6\% \\
alt & 53.1\% ± 5.7\% & 57.0\% ± 3.7\% & 41.6\% ± 8.6\% & 53.8\% ± 2.5\% & 22.7\% ± 4.1\% & 20.1\% ± 9.5\% \\
am & 21.2\% ± 2.6\% & 22.1\% ± 2.2\% & 18.7\% ± 8.6\% & 20.6\% ± 1.5\% & 21.2\% ± 2.0\% & 20.7\% ± 9.1\% \\
ami & 14.9\% ± 9.1\% & 20.2\% ± 8.0\% & 16.5\% ± 8.1\% & 25.0\% ± 4.7\% & 0.6\% ± 1.4\% & 0.0\% ± 0.0\% \\
an & 71.3\% ± 2.5\% & 69.4\% ± 2.8\% & 66.7\% ± 3.7\% & 63.6\% ± 2.4\% & 26.8\% ± 2.5\% & 29.0\% ± 2.8\% \\
ang & 55.2\% ± 3.3\% & 59.1\% ± 3.1\% & 45.3\% ± 2.8\% & 48.9\% ± 3.0\% & 8.6\% ± 6.4\% & 10.8\% ± 10.1\% \\
anp & 57.3\% ± 4.7\% & 57.2\% ± 4.2\% & 54.2\% ± 3.4\% & 50.5\% ± 3.0\% & 24.6\% ± 4.1\% & 25.5\% ± 4.4\% \\
ar & 68.8\% ± 6.1\% & 65.8\% ± 4.1\% & 63.9\% ± 5.1\% & 59.6\% ± 1.8\% & 25.4\% ± 3.5\% & 27.5\% ± 4.1\% \\
ary & 59.0\% ± 4.5\% & 59.5\% ± 2.8\% & 50.9\% ± 5.0\% & 50.4\% ± 3.0\% & 17.5\% ± 3.4\% & 22.8\% ± 5.1\% \\
arz & 71.9\% ± 6.4\% & 71.9\% ± 5.4\% & 57.5\% ± 5.4\% & 60.4\% ± 5.9\% & 51.5\% ± 2.2\% & 52.9\% ± 2.3\% \\
as & 62.1\% ± 2.5\% & 59.2\% ± 2.9\% & 58.2\% ± 3.2\% & 59.5\% ± 1.7\% & 20.5\% ± 8.8\% & 22.8\% ± 7.7\% \\
ast & 73.3\% ± 2.9\% & 71.9\% ± 3.6\% & 69.5\% ± 4.0\% & 64.4\% ± 2.5\% & 31.6\% ± 6.6\% & 37.1\% ± 3.1\% \\
av & 36.5\% ± 11.3\% & 45.1\% ± 5.0\% & 31.0\% ± 11.8\% & 40.5\% ± 8.6\% & 8.0\% ± 6.5\% & 7.6\% ± 8.4\% \\
avk & 39.7\% ± 6.6\% & 44.7\% ± 3.4\% & 10.0\% ± 7.8\% & 46.8\% ± 2.9\% & 33.5\% ± 3.0\% & 37.7\% ± 2.7\% \\
awa & 67.4\% ± 6.3\% & 68.8\% ± 4.9\% & 63.6\% ± 6.7\% & 67.0\% ± 4.5\% & 31.1\% ± 15.8\% & 35.1\% ± 16.0\% \\
ay & 33.2\% ± 5.3\% & 37.7\% ± 3.7\% & 13.4\% ± 6.4\% & 36.1\% ± 2.8\% & 11.4\% ± 2.7\% & 24.4\% ± 3.5\% \\
az & 71.5\% ± 8.8\% & 69.0\% ± 4.4\% & 62.2\% ± 7.0\% & 64.2\% ± 3.5\% & 31.1\% ± 1.7\% & 31.6\% ± 3.1\% \\
azb & 59.9\% ± 5.1\% & 56.3\% ± 10.0\% & 43.4\% ± 7.6\% & 48.3\% ± 6.0\% & 11.4\% ± 7.5\% & 25.2\% ± 5.3\% \\
ba & 65.8\% ± 6.1\% & 69.5\% ± 4.0\% & 63.2\% ± 3.8\% & 67.0\% ± 2.9\% & 15.0\% ± 6.9\% & 26.6\% ± 5.5\% \\
ban & 60.5\% ± 2.8\% & 65.3\% ± 2.1\% & 44.7\% ± 11.6\% & 61.1\% ± 3.7\% & 26.7\% ± 3.9\% & 31.8\% ± 3.4\% \\
bar & 61.0\% ± 4.3\% & 65.7\% ± 3.5\% & 53.8\% ± 4.6\% & 53.7\% ± 3.2\% & 13.1\% ± 6.4\% & 20.9\% ± 3.1\% \\
bcl & 64.3\% ± 5.4\% & 70.1\% ± 3.5\% & 54.0\% ± 16.0\% & 65.2\% ± 4.6\% & 18.4\% ± 12.5\% & 27.3\% ± 3.3\% \\
be & 65.5\% ± 6.2\% & 62.5\% ± 3.7\% & 61.1\% ± 3.9\% & 50.0\% ± 2.8\% & 31.3\% ± 3.1\% & 34.4\% ± 2.2\% \\
bg & 77.0\% ± 4.2\% & 74.2\% ± 3.7\% & 72.4\% ± 4.2\% & 65.8\% ± 2.8\% & 30.2\% ± 2.1\% & 31.6\% ± 3.1\% \\
bjn & 68.3\% ± 2.2\% & 68.3\% ± 2.6\% & 66.6\% ± 4.2\% & 69.9\% ± 2.3\% & 27.5\% ± 5.3\% & 33.0\% ± 3.5\% \\
blk & 16.4\% ± 5.3\% & 21.4\% ± 2.6\% & 13.7\% ± 2.5\% & 14.7\% ± 2.5\% & 1.4\% ± 3.1\% & 0.0\% ± 0.0\% \\
bn & 62.9\% ± 6.2\% & 64.7\% ± 2.9\% & 55.7\% ± 7.7\% & 62.8\% ± 2.2\% & 27.8\% ± 4.9\% & 29.4\% ± 3.0\% \\
bo & 0.3\% ± 0.4\% & 0.9\% ± 0.8\% & 3.2\% ± 1.6\% & 3.6\% ± 1.8\% & 0.0\% ± 0.1\% & 0.0\% ± 0.0\% \\
bpy & 60.9\% ± 14.7\% & 62.8\% ± 7.0\% & 59.2\% ± 11.5\% & 67.6\% ± 1.5\% & 35.3\% ± 4.1\% & 37.0\% ± 3.9\% \\
br & 68.0\% ± 3.6\% & 67.9\% ± 2.0\% & 58.4\% ± 5.2\% & 63.4\% ± 2.2\% & 27.0\% ± 3.4\% & 28.5\% ± 3.1\% \\
bs & 69.6\% ± 5.4\% & 69.0\% ± 2.9\% & 61.6\% ± 5.2\% & 66.2\% ± 2.2\% & 36.0\% ± 2.6\% & 36.0\% ± 3.2\% \\
bxr & 56.4\% ± 4.6\% & 55.9\% ± 7.1\% & 25.4\% ± 13.1\% & 49.1\% ± 4.5\% & 19.8\% ± 2.7\% & 21.4\% ± 9.5\% \\
ca & 73.9\% ± 4.6\% & 71.3\% ± 4.3\% & 69.8\% ± 3.9\% & 66.8\% ± 2.7\% & 30.6\% ± 4.5\% & 30.5\% ± 3.6\% \\
cdo & 34.4\% ± 6.4\% & 41.0\% ± 3.5\% & 45.2\% ± 6.6\% & 42.3\% ± 10.2\% & 6.2\% ± 8.1\% & 0.8\% ± 3.1\% \\
ce & 49.4\% ± 6.7\% & 55.1\% ± 4.1\% & 17.7\% ± 6.6\% & 45.2\% ± 3.5\% & 19.4\% ± 3.7\% & 24.7\% ± 4.9\% \\
ceb & 69.2\% ± 7.7\% & 70.7\% ± 5.5\% & 43.9\% ± 6.6\% & 61.0\% ± 7.0\% & 44.5\% ± 2.8\% & 44.9\% ± 2.6\% \\
ckb & 35.8\% ± 9.1\% & 27.9\% ± 7.9\% & 46.3\% ± 4.8\% & 48.4\% ± 3.4\% & 1.1\% ± 2.3\% & 18.4\% ± 8.4\% \\
co & 69.6\% ± 6.9\% & 64.3\% ± 4.3\% & 64.3\% ± 8.3\% & 59.3\% ± 5.6\% & 26.0\% ± 11.7\% & 33.5\% ± 4.5\% \\
crh & 63.5\% ± 4.9\% & 63.2\% ± 3.9\% & 54.3\% ± 4.2\% & 59.2\% ± 3.1\% & 21.4\% ± 3.2\% & 28.7\% ± 3.6\% \\
cs & 71.3\% ± 3.7\% & 68.2\% ± 4.1\% & 68.9\% ± 3.3\% & 64.0\% ± 3.1\% & 29.8\% ± 2.4\% & 31.2\% ± 3.6\% \\
csb & 53.9\% ± 7.0\% & 55.2\% ± 3.9\% & 45.4\% ± 6.3\% & 45.1\% ± 5.5\% & 14.3\% ± 7.0\% & 21.1\% ± 3.0\% \\
cv & 37.4\% ± 5.7\% & 41.8\% ± 3.2\% & 23.0\% ± 10.9\% & 58.3\% ± 3.5\% & 11.6\% ± 2.9\% & 19.5\% ± 5.3\% \\
cy & 90.3\% ± 1.9\% & 90.6\% ± 1.3\% & 67.1\% ± 13.0\% & 89.2\% ± 1.9\% & 49.4\% ± 2.6\% & 49.4\% ± 2.2\% \\
da & 78.9\% ± 2.7\% & 77.1\% ± 2.3\% & 75.2\% ± 2.3\% & 74.2\% ± 2.0\% & 32.3\% ± 3.9\% & 33.1\% ± 2.4\% \\
dag & 29.4\% ± 11.3\% & 36.4\% ± 6.0\% & 22.1\% ± 11.5\% & 43.8\% ± 1.6\% & 8.2\% ± 4.5\% & 6.4\% ± 9.4\% \\
de & 80.3\% ± 2.2\% & 76.6\% ± 2.4\% & 79.2\% ± 2.1\% & 70.1\% ± 2.4\% & 28.5\% ± 3.1\% & 30.5\% ± 3.6\% \\
din & 25.9\% ± 8.6\% & 27.4\% ± 7.1\% & 14.2\% ± 6.2\% & 23.3\% ± 5.7\% & 7.0\% ± 11.3\% & 7.9\% ± 12.6\% \\
diq & 43.4\% ± 4.1\% & 48.2\% ± 2.9\% & 31.8\% ± 11.1\% & 41.5\% ± 4.7\% & 13.7\% ± 6.6\% & 14.8\% ± 8.8\% \\
dsb & 61.7\% ± 3.2\% & 61.4\% ± 2.1\% & 37.9\% ± 18.0\% & 55.4\% ± 2.9\% & 24.1\% ± 3.9\% & 31.0\% ± 4.1\% \\
dty & 65.9\% ± 4.0\% & 63.1\% ± 4.4\% & 63.8\% ± 2.5\% & 62.4\% ± 3.1\% & 24.6\% ± 2.0\% & 27.1\% ± 1.6\% \\
dv & 7.2\% ± 6.0\% & 8.5\% ± 6.1\% & 18.7\% ± 8.6\% & 26.3\% ± 4.2\% & 0.1\% ± 0.2\% & 0.2\% ± 0.8\% \\
dz & 1.4\% ± 1.2\% & 1.9\% ± 1.1\% & 5.6\% ± 1.5\% & 5.9\% ± 1.4\% & 0.0\% ± 0.0\% & 0.0\% ± 0.0\% \\
el & 71.6\% ± 4.1\% & 69.2\% ± 2.4\% & 68.5\% ± 4.3\% & 58.0\% ± 3.3\% & 27.3\% ± 3.2\% & 29.3\% ± 3.3\% \\
en & 79.5\% ± 3.9\% & 78.5\% ± 2.7\% & 76.4\% ± 4.4\% & 71.4\% ± 1.7\% & 28.1\% ± 3.6\% & 30.6\% ± 1.9\% \\
eo & 73.6\% ± 5.2\% & 70.6\% ± 4.7\% & 67.9\% ± 5.0\% & 68.5\% ± 2.2\% & 30.6\% ± 3.5\% & 30.6\% ± 3.7\% \\
es & 76.6\% ± 2.6\% & 74.1\% ± 2.1\% & 72.5\% ± 4.7\% & 67.9\% ± 2.7\% & 29.4\% ± 2.6\% & 27.7\% ± 4.4\% \\
et & 73.5\% ± 3.6\% & 71.8\% ± 4.0\% & 67.8\% ± 5.0\% & 65.6\% ± 2.1\% & 30.7\% ± 2.3\% & 29.1\% ± 2.4\% \\
eu & 68.8\% ± 2.9\% & 63.1\% ± 5.2\% & 62.4\% ± 4.1\% & 63.8\% ± 3.4\% & 27.7\% ± 3.9\% & 26.9\% ± 3.3\% \\
ext & 68.2\% ± 4.4\% & 70.0\% ± 4.1\% & 65.1\% ± 4.5\% & 58.6\% ± 4.0\% & 21.7\% ± 4.1\% & 27.2\% ± 2.4\% \\
fa & 68.5\% ± 3.0\% & 66.8\% ± 1.6\% & 62.0\% ± 3.6\% & 62.9\% ± 2.0\% & 32.4\% ± 3.4\% & 32.8\% ± 2.5\% \\
fat & 30.3\% ± 5.0\% & 36.6\% ± 3.5\% & 24.1\% ± 7.7\% & 38.0\% ± 3.5\% & 1.0\% ± 2.4\% & 8.4\% ± 9.2\% \\
ff & 20.4\% ± 8.8\% & 26.0\% ± 5.2\% & 14.9\% ± 4.6\% & 31.6\% ± 2.9\% & 6.3\% ± 5.6\% & 4.5\% ± 8.8\% \\
fi & 69.4\% ± 5.6\% & 69.8\% ± 5.1\% & 62.3\% ± 4.5\% & 63.7\% ± 1.9\% & 29.7\% ± 3.4\% & 28.2\% ± 4.2\% \\
fo & 65.7\% ± 2.0\% & 67.7\% ± 2.9\% & 63.8\% ± 4.1\% & 66.4\% ± 3.3\% & 24.4\% ± 4.8\% & 27.6\% ± 4.8\% \\
\hline
\end{tabular}
\caption{The F1-scores on MultiWikiQA for languages 1-66, sorted alphabetically.}
\label{tab:num-samples-in-small-subsets}
\end{table*}

\begin{table*}[h]
\centering
\scriptsize
\begin{tabular}{crrrrrr}
\hline
\textbf{Language} & \textbf{Mistral Base} & \textbf{Mistral Instruct} & \textbf{Llama Base} & \textbf{Llama Instruct} & \textbf{XLM-RoBERTa} & \textbf{Multi-E5} \\
\hline
fr & 74.1\% ± 4.7\% & 71.6\% ± 1.6\% & 71.5\% ± 3.7\% & 66.5\% ± 1.4\% & 28.0\% ± 2.4\% & 29.1\% ± 4.7\% \\
frp & 62.5\% ± 3.7\% & 59.2\% ± 3.8\% & 54.8\% ± 2.8\% & 45.1\% ± 3.3\% & 14.4\% ± 6.4\% & 26.2\% ± 3.5\% \\
frr & 53.9\% ± 4.2\% & 58.7\% ± 2.9\% & 49.1\% ± 5.1\% & 53.5\% ± 2.5\% & 17.5\% ± 2.5\% & 23.5\% ± 3.0\% \\
fur & 63.0\% ± 5.6\% & 64.8\% ± 4.6\% & 54.0\% ± 9.0\% & 60.7\% ± 2.5\% & 7.8\% ± 4.2\% & 25.8\% ± 2.6\% \\
fy & 68.0\% ± 4.0\% & 69.5\% ± 2.4\% & 68.6\% ± 6.2\% & 73.2\% ± 2.1\% & 27.2\% ± 3.1\% & 29.1\% ± 3.3\% \\
ga & 68.5\% ± 6.0\% & 69.7\% ± 3.3\% & 56.6\% ± 7.7\% & 62.9\% ± 4.9\% & 24.2\% ± 2.7\% & 26.9\% ± 4.3\% \\
gag & 65.2\% ± 2.5\% & 66.1\% ± 1.4\% & 57.1\% ± 6.4\% & 57.8\% ± 1.7\% & 27.3\% ± 4.5\% & 33.5\% ± 3.4\% \\
gcr & 70.7\% ± 5.4\% & 72.8\% ± 3.2\% & 61.0\% ± 4.1\% & 59.2\% ± 6.7\% & 26.7\% ± 2.3\% & 34.8\% ± 2.3\% \\
gd & 70.3\% ± 3.6\% & 66.8\% ± 3.2\% & 58.0\% ± 7.2\% & 64.7\% ± 1.7\% & 31.2\% ± 3.1\% & 30.4\% ± 2.5\% \\
gl & 78.4\% ± 3.9\% & 75.4\% ± 4.7\% & 71.9\% ± 4.5\% & 66.4\% ± 2.5\% & 31.2\% ± 2.5\% & 29.6\% ± 5.1\% \\
glk & 39.9\% ± 9.0\% & 37.6\% ± 6.5\% & 33.9\% ± 5.5\% & 33.8\% ± 3.4\% & 13.2\% ± 4.2\% & 14.2\% ± 12.7\% \\
gn & 21.4\% ± 6.2\% & 26.6\% ± 4.2\% & 22.3\% ± 9.4\% & 37.0\% ± 2.3\% & 1.8\% ± 2.2\% & 1.7\% ± 6.6\% \\
gom & 42.5\% ± 2.8\% & 47.1\% ± 2.6\% & 41.3\% ± 5.3\% & 47.7\% ± 2.5\% & 8.4\% ± 5.1\% & 6.0\% ± 8.6\% \\
gor & 61.4\% ± 3.9\% & 64.0\% ± 2.8\% & 54.1\% ± 5.5\% & 60.9\% ± 3.5\% & 31.3\% ± 2.2\% & 33.8\% ± 2.1\% \\
gpe & 75.0\% ± 2.5\% & 74.5\% ± 1.9\% & 71.0\% ± 3.8\% & 69.3\% ± 2.5\% & 32.1\% ± 2.9\% & 33.5\% ± 3.7\% \\
gu & 61.2\% ± 4.2\% & 62.3\% ± 3.8\% & 57.4\% ± 1.6\% & 53.9\% ± 3.4\% & 28.5\% ± 3.1\% & 27.2\% ± 5.0\% \\
guc & 19.6\% ± 6.2\% & 22.7\% ± 5.0\% & 8.7\% ± 3.6\% & 18.5\% ± 3.1\% & 0.6\% ± 2.3\% & 0.0\% ± 0.1\% \\
gur & 37.2\% ± 5.7\% & 39.7\% ± 3.6\% & 20.9\% ± 9.6\% & 38.5\% ± 1.8\% & 10.1\% ± 6.1\% & 16.4\% ± 7.0\% \\
guw & 17.5\% ± 9.2\% & 24.8\% ± 4.8\% & 24.1\% ± 5.0\% & 32.8\% ± 2.5\% & 0.2\% ± 0.7\% & 5.0\% ± 8.0\% \\
gv & 52.2\% ± 4.5\% & 53.8\% ± 2.5\% & 43.0\% ± 8.6\% & 51.8\% ± 2.0\% & 13.8\% ± 3.3\% & 15.4\% ± 7.5\% \\
ha & 52.4\% ± 4.9\% & 55.6\% ± 2.6\% & 57.6\% ± 3.4\% & 58.3\% ± 2.3\% & 25.4\% ± 4.1\% & 25.9\% ± 5.2\% \\
hak & 37.4\% ± 8.6\% & 39.5\% ± 7.5\% & 40.8\% ± 10.6\% & 44.2\% ± 6.8\% & 5.9\% ± 7.5\% & 6.2\% ± 9.8\% \\
haw & 38.5\% ± 11.3\% & 46.8\% ± 7.8\% & 24.0\% ± 8.7\% & 40.9\% ± 6.4\% & 4.5\% ± 6.7\% & 8.4\% ± 12.7\% \\
he & 73.2\% ± 3.8\% & 69.2\% ± 3.1\% & 63.7\% ± 3.9\% & 63.6\% ± 2.8\% & 24.2\% ± 3.4\% & 26.2\% ± 4.1\% \\
hi & 67.6\% ± 5.3\% & 67.4\% ± 3.5\% & 65.5\% ± 8.0\% & 65.9\% ± 2.1\% & 28.5\% ± 1.5\% & 28.5\% ± 2.6\% \\
hif & 70.6\% ± 4.3\% & 67.2\% ± 2.0\% & 60.8\% ± 3.9\% & 66.3\% ± 2.8\% & 25.0\% ± 3.5\% & 26.3\% ± 3.6\% \\
hr & 64.6\% ± 5.0\% & 64.7\% ± 2.5\% & 64.3\% ± 2.8\% & 60.9\% ± 2.4\% & 28.1\% ± 2.7\% & 30.2\% ± 2.6\% \\
hsb & 60.2\% ± 4.1\% & 63.9\% ± 2.4\% & 47.0\% ± 11.7\% & 52.8\% ± 3.5\% & 28.5\% ± 4.7\% & 33.9\% ± 3.9\% \\
ht & 79.3\% ± 4.0\% & 76.9\% ± 3.2\% & 67.1\% ± 2.9\% & 70.5\% ± 2.3\% & 36.3\% ± 1.8\% & 44.2\% ± 1.4\% \\
hu & 66.6\% ± 7.5\% & 65.3\% ± 5.9\% & 63.2\% ± 5.7\% & 63.2\% ± 3.8\% & 25.8\% ± 3.0\% & 28.4\% ± 1.7\% \\
hy & 64.0\% ± 10.4\% & 62.6\% ± 6.9\% & 47.6\% ± 5.1\% & 48.9\% ± 3.2\% & 32.6\% ± 2.0\% & 33.1\% ± 2.6\% \\
hyw & 56.3\% ± 7.9\% & 54.0\% ± 8.0\% & 44.5\% ± 2.6\% & 44.8\% ± 3.0\% & 29.0\% ± 2.8\% & 29.8\% ± 2.0\% \\
ia & 72.1\% ± 2.9\% & 69.1\% ± 3.4\% & 63.8\% ± 2.3\% & 60.4\% ± 2.7\% & 28.7\% ± 2.3\% & 33.6\% ± 2.5\% \\
id & 77.9\% ± 3.9\% & 75.0\% ± 3.5\% & 75.0\% ± 3.4\% & 71.0\% ± 2.5\% & 34.9\% ± 3.0\% & 35.7\% ± 3.2\% \\
ie & 76.3\% ± 4.3\% & 75.6\% ± 4.4\% & 73.4\% ± 3.3\% & 72.7\% ± 2.5\% & 35.1\% ± 1.8\% & 36.0\% ± 2.8\% \\
ig & 45.2\% ± 5.6\% & 53.3\% ± 3.4\% & 49.0\% ± 6.9\% & 60.7\% ± 2.1\% & 2.5\% ± 2.7\% & 18.4\% ± 12.5\% \\
ilo & 56.2\% ± 4.8\% & 62.2\% ± 2.4\% & 37.8\% ± 16.2\% & 61.4\% ± 2.6\% & 9.0\% ± 6.0\% & 24.0\% ± 5.9\% \\
inh & 33.7\% ± 6.9\% & 39.2\% ± 6.1\% & 18.3\% ± 7.3\% & 29.3\% ± 4.3\% & 13.6\% ± 8.6\% & 11.2\% ± 14.5\% \\
io & 68.5\% ± 5.2\% & 64.2\% ± 5.0\% & 68.1\% ± 3.8\% & 66.7\% ± 2.3\% & 39.3\% ± 3.4\% & 42.2\% ± 1.8\% \\
is & 70.7\% ± 2.8\% & 68.6\% ± 2.5\% & 66.0\% ± 4.1\% & 63.5\% ± 1.6\% & 27.7\% ± 2.0\% & 29.5\% ± 3.5\% \\
it & 76.3\% ± 3.5\% & 73.0\% ± 1.9\% & 72.3\% ± 4.4\% & 65.0\% ± 1.5\% & 25.0\% ± 3.2\% & 25.8\% ± 3.1\% \\
ja & 50.0\% ± 3.4\% & 47.4\% ± 4.3\% & 48.1\% ± 4.3\% & 41.5\% ± 3.5\% & 14.9\% ± 3.6\% & 17.8\% ± 3.6\% \\
jam & 54.5\% ± 6.2\% & 54.6\% ± 5.4\% & 39.1\% ± 4.9\% & 38.5\% ± 6.8\% & 3.9\% ± 6.4\% & 19.7\% ± 6.7\% \\
ka & 66.4\% ± 7.0\% & 62.4\% ± 3.9\% & 51.7\% ± 2.2\% & 52.7\% ± 1.9\% & 32.7\% ± 3.8\% & 31.9\% ± 2.1\% \\
kaa & 58.1\% ± 5.9\% & 57.1\% ± 4.6\% & 46.2\% ± 5.2\% & 48.5\% ± 3.1\% & 13.9\% ± 4.9\% & 17.1\% ± 8.7\% \\
kab & 22.7\% ± 7.3\% & 28.5\% ± 6.3\% & 14.9\% ± 4.4\% & 30.3\% ± 2.4\% & 1.5\% ± 2.3\% & 3.5\% ± 6.4\% \\
kbd & 32.4\% ± 5.8\% & 36.9\% ± 5.0\% & 27.7\% ± 7.3\% & 29.1\% ± 3.6\% & 5.3\% ± 7.5\% & 0.0\% ± 0.0\% \\
kbp & 13.4\% ± 11.7\% & 18.9\% ± 13.3\% & 13.0\% ± 7.1\% & 30.8\% ± 3.7\% & 3.0\% ± 4.6\% & 6.6\% ± 12.0\% \\
kk & 69.6\% ± 8.1\% & 63.6\% ± 9.0\% & 60.6\% ± 7.0\% & 59.4\% ± 5.5\% & 32.7\% ± 3.0\% & 32.4\% ± 5.5\% \\
km & 7.7\% ± 5.0\% & 7.4\% ± 3.8\% & 26.4\% ± 8.3\% & 24.6\% ± 3.4\% & 25.3\% ± 3.4\% & 27.7\% ± 3.6\% \\
kn & 66.7\% ± 2.8\% & 63.4\% ± 1.9\% & 53.8\% ± 3.3\% & 54.6\% ± 2.1\% & 29.2\% ± 6.6\% & 26.6\% ± 8.5\% \\
ko & 60.8\% ± 5.5\% & 57.0\% ± 4.3\% & 56.4\% ± 5.1\% & 51.8\% ± 2.7\% & 18.0\% ± 12.1\% & 17.6\% ± 10.5\% \\
koi & 37.1\% ± 9.8\% & 47.5\% ± 3.0\% & 7.5\% ± 6.3\% & 45.8\% ± 2.2\% & 17.9\% ± 8.8\% & 17.6\% ± 13.1\% \\
krc & 59.9\% ± 3.5\% & 58.4\% ± 3.9\% & 48.7\% ± 3.3\% & 52.2\% ± 2.0\% & 23.1\% ± 8.5\% & 26.9\% ± 11.6\% \\
ks & 44.8\% ± 4.4\% & 45.8\% ± 3.5\% & 38.7\% ± 5.6\% & 42.2\% ± 5.9\% & 11.7\% ± 7.9\% & 25.2\% ± 11.7\% \\
ku & 62.5\% ± 5.0\% & 67.2\% ± 2.6\% & 46.2\% ± 5.3\% & 51.2\% ± 2.8\% & 21.5\% ± 8.5\% & 22.7\% ± 4.7\% \\
kv & 48.5\% ± 6.7\% & 50.7\% ± 6.9\% & 24.9\% ± 14.0\% & 50.6\% ± 3.4\% & 19.5\% ± 5.7\% & 24.0\% ± 5.2\% \\
kw & 41.7\% ± 11.3\% & 47.5\% ± 5.1\% & 24.9\% ± 12.1\% & 51.8\% ± 3.5\% & 13.0\% ± 4.0\% & 15.5\% ± 4.8\% \\
ky & 59.8\% ± 9.9\% & 58.2\% ± 6.0\% & 56.7\% ± 2.9\% & 60.0\% ± 3.2\% & 31.6\% ± 2.1\% & 30.2\% ± 2.2\% \\
la & 67.5\% ± 5.8\% & 67.3\% ± 4.7\% & 58.6\% ± 4.4\% & 63.1\% ± 2.6\% & 25.5\% ± 2.3\% & 28.6\% ± 3.7\% \\
lad & 70.2\% ± 2.6\% & 66.8\% ± 2.8\% & 63.5\% ± 3.0\% & 55.1\% ± 2.8\% & 26.8\% ± 3.5\% & 32.1\% ± 3.2\% \\
lb & 71.0\% ± 3.8\% & 73.0\% ± 6.3\% & 65.4\% ± 4.0\% & 68.3\% ± 2.6\% & 16.7\% ± 4.5\% & 31.3\% ± 4.1\% \\
lez & 48.0\% ± 5.0\% & 52.4\% ± 4.0\% & 42.1\% ± 4.2\% & 49.3\% ± 2.5\% & 15.7\% ± 6.4\% & 19.4\% ± 7.1\% \\
lfn & 71.0\% ± 2.4\% & 67.1\% ± 1.8\% & 62.4\% ± 3.7\% & 59.8\% ± 2.2\% & 23.9\% ± 4.2\% & 27.3\% ± 4.9\% \\
lg & 28.4\% ± 11.8\% & 42.4\% ± 6.4\% & 20.0\% ± 7.1\% & 39.7\% ± 1.6\% & 3.6\% ± 4.7\% & 14.3\% ± 9.0\% \\
\hline
\end{tabular}
\caption{The F1-scores on MultiWikiQA for languages 66-131, sorted alphabetically.}
\label{tab:num-samples-in-small-subsets}
\end{table*}

\begin{table*}[h]
\centering
\scriptsize
\begin{tabular}{crrrrrr}
\hline
\textbf{Language} & \textbf{Mistral Base} & \textbf{Mistral Instruct} & \textbf{Llama Base} & \textbf{Llama Instruct} & \textbf{XLM-RoBERTa} & \textbf{Multi-E5} \\
\hline
li & 64.6\% ± 3.3\% & 66.2\% ± 2.9\% & 60.9\% ± 2.9\% & 62.3\% ± 1.8\% & 18.2\% ± 8.7\% & 24.3\% ± 4.5\% \\
lij & 66.9\% ± 2.6\% & 64.4\% ± 3.9\% & 58.2\% ± 3.1\% & 55.1\% ± 1.8\% & 9.6\% ± 7.5\% & 19.7\% ± 6.4\% \\
lld & 51.9\% ± 9.8\% & 54.0\% ± 1.9\% & 39.3\% ± 14.9\% & 51.7\% ± 3.3\% & 8.8\% ± 6.8\% & 21.8\% ± 9.5\% \\
lmo & 64.9\% ± 4.2\% & 66.0\% ± 4.7\% & 54.6\% ± 11.3\% & 57.9\% ± 2.3\% & 11.4\% ± 6.2\% & 22.4\% ± 4.4\% \\
ln & 32.9\% ± 19.6\% & 44.8\% ± 12.0\% & 26.8\% ± 12.0\% & 46.3\% ± 2.9\% & 4.3\% ± 3.6\% & 25.5\% ± 7.2\% \\
lo & 13.0\% ± 5.7\% & 11.9\% ± 3.9\% & 9.8\% ± 7.7\% & 21.5\% ± 2.5\% & 23.5\% ± 4.0\% & 24.8\% ± 8.1\% \\
lt & 70.1\% ± 5.2\% & 67.9\% ± 4.7\% & 68.1\% ± 3.0\% & 64.3\% ± 2.6\% & 33.9\% ± 2.1\% & 34.6\% ± 4.4\% \\
ltg & 61.1\% ± 5.4\% & 64.4\% ± 3.9\% & 38.7\% ± 6.8\% & 52.2\% ± 6.6\% & 19.0\% ± 13.0\% & 24.9\% ± 13.2\% \\
lv & 67.3\% ± 5.9\% & 70.0\% ± 4.1\% & 60.6\% ± 5.7\% & 65.7\% ± 1.6\% & 33.1\% ± 2.5\% & 30.3\% ± 13.0\% \\
mad & 56.8\% ± 5.4\% & 59.5\% ± 3.5\% & 58.1\% ± 4.6\% & 59.4\% ± 2.9\% & 11.9\% ± 8.8\% & 23.4\% ± 5.6\% \\
mai & 63.3\% ± 5.8\% & 64.0\% ± 4.4\% & 59.3\% ± 3.7\% & 60.2\% ± 3.0\% & 29.4\% ± 2.6\% & 29.4\% ± 5.1\% \\
mdf & 36.9\% ± 10.0\% & 46.1\% ± 3.1\% & 14.6\% ± 6.8\% & 42.6\% ± 2.3\% & 22.6\% ± 2.7\% & 22.7\% ± 3.0\% \\
mg & 48.8\% ± 7.1\% & 54.3\% ± 4.0\% & 21.5\% ± 9.5\% & 45.0\% ± 2.5\% & 34.7\% ± 14.0\% & 35.0\% ± 16.3\% \\
mhr & 48.9\% ± 5.7\% & 53.2\% ± 3.6\% & 31.3\% ± 10.5\% & 53.9\% ± 3.6\% & 25.3\% ± 6.1\% & 30.1\% ± 4.9\% \\
mi & 64.8\% ± 3.9\% & 66.4\% ± 3.9\% & 54.3\% ± 11.4\% & 58.4\% ± 5.3\% & 16.7\% ± 11.0\% & 34.9\% ± 6.2\% \\
min & 66.5\% ± 3.3\% & 67.1\% ± 3.0\% & 64.7\% ± 4.1\% & 62.0\% ± 2.1\% & 26.2\% ± 2.8\% & 29.1\% ± 3.0\% \\
mk & 74.4\% ± 3.9\% & 72.3\% ± 2.1\% & 70.6\% ± 3.7\% & 73.2\% ± 1.8\% & 31.3\% ± 3.1\% & 31.5\% ± 2.1\% \\
ml & 64.9\% ± 1.5\% & 64.9\% ± 2.8\% & 52.6\% ± 1.9\% & 52.5\% ± 2.4\% & 29.9\% ± 3.5\% & 30.8\% ± 2.6\% \\
mn & 63.3\% ± 6.0\% & 64.0\% ± 4.5\% & 52.6\% ± 11.2\% & 57.2\% ± 2.4\% & 28.9\% ± 2.4\% & 29.1\% ± 2.3\% \\
mni & 1.1\% ± 0.7\% & 1.0\% ± 0.6\% & 2.0\% ± 0.7\% & 2.7\% ± 1.5\% & 0.0\% ± 0.0\% & 0.0\% ± 0.0\% \\
mnw & 4.7\% ± 4.1\% & 7.7\% ± 3.5\% & 12.8\% ± 4.5\% & 21.2\% ± 4.3\% & 0.1\% ± 0.3\% & 0.0\% ± 0.0\% \\
mr & 71.8\% ± 3.8\% & 71.8\% ± 1.8\% & 65.2\% ± 6.1\% & 72.8\% ± 1.8\% & 33.3\% ± 3.3\% & 32.2\% ± 5.8\% \\
mrj & 44.0\% ± 9.7\% & 49.9\% ± 5.7\% & 37.8\% ± 10.1\% & 52.8\% ± 2.6\% & 17.5\% ± 3.3\% & 23.6\% ± 9.2\% \\
ms & 81.6\% ± 2.2\% & 77.1\% ± 2.2\% & 76.7\% ± 1.6\% & 76.1\% ± 2.3\% & 33.3\% ± 3.5\% & 32.3\% ± 3.1\% \\
mt & 61.7\% ± 4.7\% & 61.5\% ± 3.6\% & 56.1\% ± 3.6\% & 59.9\% ± 1.7\% & 1.9\% ± 2.8\% & 18.2\% ± 9.3\% \\
mwl & 76.1\% ± 4.3\% & 73.9\% ± 1.9\% & 67.4\% ± 7.6\% & 60.6\% ± 2.4\% & 26.3\% ± 4.4\% & 31.3\% ± 5.8\% \\
my & 26.4\% ± 7.8\% & 29.8\% ± 5.8\% & 19.2\% ± 5.9\% & 23.0\% ± 5.1\% & 24.2\% ± 5.6\% & 23.0\% ± 4.4\% \\
myv & 39.7\% ± 4.0\% & 43.2\% ± 3.2\% & 27.4\% ± 8.3\% & 43.2\% ± 2.5\% & 13.4\% ± 5.9\% & 16.2\% ± 10.7\% \\
mzn & 56.3\% ± 7.2\% & 57.5\% ± 5.4\% & 50.9\% ± 6.3\% & 51.0\% ± 3.5\% & 19.7\% ± 3.2\% & 23.8\% ± 2.1\% \\
nap & 66.8\% ± 3.4\% & 67.1\% ± 3.3\% & 57.1\% ± 4.2\% & 58.3\% ± 2.1\% & 8.9\% ± 8.9\% & 24.9\% ± 5.0\% \\
nds & 72.5\% ± 3.1\% & 71.5\% ± 3.4\% & 66.6\% ± 4.2\% & 69.9\% ± 2.4\% & 25.8\% ± 5.4\% & 31.7\% ± 4.4\% \\
ne & 62.8\% ± 4.7\% & 62.4\% ± 3.5\% & 65.0\% ± 2.8\% & 65.4\% ± 2.4\% & 29.5\% ± 1.3\% & 29.3\% ± 3.0\% \\
new & 39.4\% ± 15.3\% & 46.1\% ± 12.1\% & 23.2\% ± 16.7\% & 50.4\% ± 10.0\% & 46.1\% ± 3.9\% & 40.7\% ± 17.5\% \\
nia & 24.2\% ± 6.9\% & 27.1\% ± 5.1\% & 15.9\% ± 5.8\% & 28.4\% ± 2.3\% & 0.5\% ± 1.5\% & 0.2\% ± 0.9\% \\
nl & 78.3\% ± 4.0\% & 78.7\% ± 2.8\% & 74.6\% ± 4.2\% & 69.0\% ± 1.7\% & 30.4\% ± 2.8\% & 31.1\% ± 3.3\% \\
nn & 77.9\% ± 4.5\% & 76.2\% ± 2.8\% & 74.6\% ± 4.7\% & 69.2\% ± 1.7\% & 32.4\% ± 3.2\% & 32.7\% ± 3.4\% \\
no & 77.1\% ± 4.3\% & 76.7\% ± 2.3\% & 74.2\% ± 3.7\% & 70.4\% ± 2.7\% & 31.6\% ± 2.8\% & 33.4\% ± 2.2\% \\
nqo & 2.5\% ± 0.9\% & 2.3\% ± 0.9\% & 2.6\% ± 0.4\% & 2.5\% ± 0.6\% & 0.0\% ± 0.1\% & 0.0\% ± 0.1\% \\
ny & 51.5\% ± 4.1\% & 56.1\% ± 3.6\% & 33.6\% ± 7.4\% & 45.6\% ± 2.4\% & 7.2\% ± 5.5\% & 24.1\% ± 4.7\% \\
oc & 65.0\% ± 4.8\% & 63.6\% ± 2.2\% & 59.2\% ± 4.0\% & 58.7\% ± 1.3\% & 23.3\% ± 2.6\% & 29.5\% ± 1.8\% \\
olo & 60.0\% ± 4.4\% & 61.9\% ± 3.1\% & 53.8\% ± 4.0\% & 55.1\% ± 3.0\% & 24.0\% ± 2.9\% & 27.4\% ± 3.6\% \\
om & 18.2\% ± 8.2\% & 27.3\% ± 4.9\% & 13.0\% ± 6.6\% & 21.7\% ± 9.1\% & 9.6\% ± 5.5\% & 9.9\% ± 9.3\% \\
or & 33.2\% ± 1.8\% & 31.1\% ± 2.1\% & 40.8\% ± 5.4\% & 41.3\% ± 1.6\% & 28.8\% ± 3.2\% & 32.3\% ± 3.7\% \\
os & 44.3\% ± 6.2\% & 48.6\% ± 4.3\% & 41.4\% ± 6.4\% & 54.1\% ± 3.6\% & 14.1\% ± 4.2\% & 11.8\% ± 13.3\% \\
pa & 73.9\% ± 3.4\% & 74.0\% ± 2.6\% & 60.0\% ± 4.6\% & 55.2\% ± 5.2\% & 27.4\% ± 4.2\% & 31.1\% ± 2.9\% \\
pam & 40.1\% ± 15.2\% & 52.3\% ± 2.4\% & 39.8\% ± 9.7\% & 54.8\% ± 2.3\% & 16.5\% ± 3.2\% & 25.1\% ± 2.2\% \\
pap & 74.0\% ± 2.0\% & 73.9\% ± 2.4\% & 68.7\% ± 4.9\% & 67.2\% ± 3.7\% & 25.4\% ± 3.3\% & 30.5\% ± 2.6\% \\
pcd & 70.6\% ± 5.3\% & 68.5\% ± 4.3\% & 63.7\% ± 4.9\% & 60.9\% ± 3.0\% & 20.7\% ± 8.5\% & 31.3\% ± 2.2\% \\
pcm & 78.3\% ± 1.7\% & 76.6\% ± 2.0\% & 73.6\% ± 2.2\% & 66.5\% ± 1.9\% & 22.4\% ± 9.9\% & 30.9\% ± 2.1\% \\
pdc & 58.2\% ± 7.8\% & 66.7\% ± 7.0\% & 51.7\% ± 5.0\% & 47.8\% ± 4.5\% & 15.1\% ± 9.6\% & 25.0\% ± 9.3\% \\
pfl & 64.3\% ± 2.0\% & 68.4\% ± 1.7\% & 55.6\% ± 4.3\% & 52.8\% ± 4.8\% & 17.8\% ± 4.2\% & 21.3\% ± 5.4\% \\
pl & 69.6\% ± 5.5\% & 64.7\% ± 3.8\% & 58.1\% ± 6.7\% & 55.8\% ± 2.6\% & 28.2\% ± 5.5\% & 32.8\% ± 4.3\% \\
pms & 66.7\% ± 3.4\% & 67.2\% ± 4.0\% & 59.1\% ± 5.1\% & 58.1\% ± 2.9\% & 15.0\% ± 5.6\% & 21.7\% ± 7.2\% \\
pnb & 61.6\% ± 3.4\% & 58.9\% ± 1.8\% & 60.1\% ± 2.4\% & 59.7\% ± 2.1\% & 19.1\% ± 5.3\% & 25.3\% ± 9.0\% \\
ps & 49.9\% ± 6.7\% & 50.9\% ± 4.2\% & 44.7\% ± 2.9\% & 41.8\% ± 4.1\% & 26.5\% ± 5.2\% & 24.8\% ± 5.8\% \\
pt-br & 79.2\% ± 5.3\% & 75.0\% ± 6.2\% & 76.3\% ± 3.9\% & 69.3\% ± 2.6\% & 28.7\% ± 3.2\% & 29.3\% ± 4.4\% \\
pt-pt & 77.7\% ± 1.5\% & 74.1\% ± 2.3\% & 74.2\% ± 3.9\% & 67.0\% ± 1.6\% & 31.0\% ± 3.5\% & 30.0\% ± 6.3\% \\
pwn & 20.4\% ± 7.4\% & 24.2\% ± 5.5\% & 15.2\% ± 5.5\% & 25.4\% ± 4.7\% & 3.1\% ± 5.2\% & 5.8\% ± 9.2\% \\
qu & 50.5\% ± 12.1\% & 56.1\% ± 3.2\% & 13.4\% ± 7.9\% & 53.6\% ± 2.3\% & 16.0\% ± 3.6\% & 26.7\% ± 4.6\% \\
rm & 60.5\% ± 4.0\% & 65.4\% ± 3.1\% & 54.2\% ± 2.5\% & 58.6\% ± 2.8\% & 17.4\% ± 2.1\% & 28.5\% ± 3.4\% \\
ro & 75.2\% ± 2.2\% & 72.9\% ± 2.6\% & 71.1\% ± 3.4\% & 68.8\% ± 2.7\% & 26.6\% ± 2.8\% & 27.4\% ± 5.0\% \\
ru & 63.0\% ± 6.1\% & 59.9\% ± 5.2\% & 58.2\% ± 5.2\% & 51.3\% ± 4.3\% & 27.3\% ± 1.8\% & 27.8\% ± 2.3\% \\
rue & 64.5\% ± 2.4\% & 62.4\% ± 2.5\% & 57.1\% ± 3.6\% & 45.8\% ± 2.3\% & 29.9\% ± 3.7\% & 33.8\% ± 3.6\% \\
rw & 44.5\% ± 3.4\% & 50.7\% ± 3.2\% & 21.1\% ± 8.0\% & 43.6\% ± 1.8\% & 7.0\% ± 3.6\% & 23.9\% ± 4.7\% \\
sa & 50.9\% ± 4.8\% & 55.8\% ± 4.5\% & 42.1\% ± 13.9\% & 48.5\% ± 2.6\% & 22.7\% ± 1.9\% & 20.3\% ± 9.1\% \\
\hline
\end{tabular}
\caption{The F1-scores on MultiWikiQA for languages 131-196, sorted alphabetically.}
\label{tab:num-samples-in-small-subsets}
\end{table*}

\begin{table*}[h]
\centering
\scriptsize
\begin{tabular}{crrrrrr}
\hline
\textbf{Language} & \textbf{Mistral Base} & \textbf{Mistral Instruct} & \textbf{Llama Base} & \textbf{Llama Instruct} & \textbf{XLM-RoBERTa} & \textbf{Multi-E5} \\
\hline
sah & 41.1\% ± 4.4\% & 44.8\% ± 3.7\% & 43.6\% ± 5.8\% & 52.0\% ± 1.9\% & 5.8\% ± 6.2\% & 7.9\% ± 12.2\% \\
sat & 5.0\% ± 3.1\% & 4.7\% ± 3.3\% & 37.3\% ± 4.1\% & 36.2\% ± 1.6\% & 0.0\% ± 0.0\% & 0.0\% ± 0.0\% \\
sc & 64.0\% ± 3.7\% & 62.8\% ± 3.2\% & 56.8\% ± 5.8\% & 54.3\% ± 2.5\% & 13.7\% ± 6.2\% & 25.1\% ± 4.4\% \\
scn & 65.4\% ± 6.2\% & 69.3\% ± 3.7\% & 54.2\% ± 5.7\% & 59.2\% ± 3.4\% & 9.2\% ± 8.2\% & 23.7\% ± 4.7\% \\
sco & 75.6\% ± 3.6\% & 75.7\% ± 2.9\% & 71.4\% ± 5.1\% & 68.4\% ± 3.1\% & 24.7\% ± 2.9\% & 30.7\% ± 4.1\% \\
sd & 45.7\% ± 5.2\% & 45.1\% ± 5.6\% & 52.6\% ± 7.3\% & 57.6\% ± 4.5\% & 29.3\% ± 4.4\% & 28.1\% ± 1.6\% \\
se & 54.5\% ± 6.6\% & 58.9\% ± 5.0\% & 29.4\% ± 10.3\% & 46.5\% ± 5.2\% & 13.7\% ± 2.7\% & 16.5\% ± 5.1\% \\
shi & 24.8\% ± 4.3\% & 27.7\% ± 2.3\% & 22.7\% ± 4.6\% & 23.2\% ± 2.8\% & 0.1\% ± 0.2\% & 0.2\% ± 0.6\% \\
shn & 3.2\% ± 3.6\% & 5.7\% ± 3.7\% & 16.4\% ± 2.9\% & 14.2\% ± 3.1\% & 0.2\% ± 0.4\% & 0.0\% ± 0.0\% \\
si & 11.0\% ± 7.7\% & 11.9\% ± 3.7\% & 37.9\% ± 4.9\% & 44.9\% ± 2.3\% & 24.7\% ± 2.1\% & 23.8\% ± 2.5\% \\
sk & 72.4\% ± 4.1\% & 70.1\% ± 3.2\% & 65.3\% ± 4.8\% & 59.5\% ± 1.8\% & 27.9\% ± 4.6\% & 28.9\% ± 3.5\% \\
skr & 49.3\% ± 2.8\% & 46.3\% ± 4.1\% & 47.5\% ± 2.6\% & 46.9\% ± 2.3\% & 13.9\% ± 6.8\% & 15.0\% ± 7.6\% \\
sl & 71.5\% ± 7.4\% & 69.8\% ± 4.7\% & 64.3\% ± 5.3\% & 64.3\% ± 2.7\% & 26.3\% ± 5.1\% & 26.2\% ± 6.4\% \\
smn & 53.7\% ± 2.7\% & 56.0\% ± 2.7\% & 21.7\% ± 13.4\% & 53.1\% ± 2.8\% & 20.1\% ± 5.2\% & 27.5\% ± 4.1\% \\
sn & 43.1\% ± 5.9\% & 49.9\% ± 5.0\% & 12.7\% ± 6.1\% & 36.1\% ± 4.9\% & 12.5\% ± 11.1\% & 21.5\% ± 10.0\% \\
so & 39.1\% ± 11.1\% & 47.7\% ± 3.0\% & 29.5\% ± 3.7\% & 42.7\% ± 3.7\% & 21.8\% ± 2.3\% & 22.1\% ± 2.0\% \\
sq & 67.8\% ± 5.5\% & 64.3\% ± 3.9\% & 60.6\% ± 5.0\% & 60.3\% ± 1.4\% & 26.5\% ± 2.0\% & 26.1\% ± 5.8\% \\
sr & 62.3\% ± 6.6\% & 60.6\% ± 4.3\% & 57.9\% ± 6.2\% & 59.9\% ± 5.3\% & 28.7\% ± 2.2\% & 26.4\% ± 11.6\% \\
ss & 36.6\% ± 6.8\% & 43.6\% ± 6.3\% & 14.4\% ± 7.8\% & 30.8\% ± 3.8\% & 5.2\% ± 5.5\% & 19.7\% ± 5.3\% \\
stq & 56.4\% ± 4.3\% & 61.6\% ± 2.8\% & 53.8\% ± 5.0\% & 59.0\% ± 2.0\% & 17.0\% ± 4.6\% & 24.2\% ± 3.3\% \\
su & 65.0\% ± 4.1\% & 69.7\% ± 1.7\% & 43.3\% ± 13.6\% & 70.0\% ± 2.9\% & 25.7\% ± 3.5\% & 28.0\% ± 2.5\% \\
sv & 78.3\% ± 2.4\% & 79.0\% ± 1.8\% & 72.8\% ± 5.1\% & 70.8\% ± 1.6\% & 33.9\% ± 2.8\% & 33.1\% ± 2.9\% \\
sw & 77.8\% ± 3.4\% & 79.4\% ± 1.1\% & 69.0\% ± 2.2\% & 72.8\% ± 1.4\% & 30.6\% ± 2.7\% & 30.0\% ± 4.2\% \\
szl & 58.3\% ± 9.0\% & 59.7\% ± 6.8\% & 50.2\% ± 9.9\% & 47.4\% ± 7.2\% & 16.3\% ± 7.6\% & 26.0\% ± 3.1\% \\
szy & 15.8\% ± 6.1\% & 24.0\% ± 3.2\% & 11.4\% ± 5.2\% & 29.4\% ± 3.3\% & 7.4\% ± 2.4\% & 2.4\% ± 5.6\% \\
ta & 61.4\% ± 6.3\% & 65.8\% ± 4.2\% & 56.0\% ± 5.3\% & 59.4\% ± 1.9\% & 34.6\% ± 2.9\% & 31.9\% ± 3.1\% \\
tay & 35.3\% ± 5.4\% & 39.1\% ± 4.6\% & 10.3\% ± 5.6\% & 35.5\% ± 3.1\% & 16.7\% ± 4.9\% & 18.6\% ± 7.1\% \\
tcy & 41.0\% ± 2.6\% & 43.5\% ± 2.8\% & 36.2\% ± 5.0\% & 37.4\% ± 3.1\% & 7.4\% ± 6.2\% & 10.6\% ± 9.5\% \\
te & 74.5\% ± 1.4\% & 71.1\% ± 2.0\% & 60.7\% ± 3.2\% & 60.0\% ± 2.2\% & 31.5\% ± 1.7\% & 27.9\% ± 12.7\% \\
tet & 66.3\% ± 4.5\% & 67.1\% ± 2.5\% & 53.1\% ± 4.9\% & 62.2\% ± 3.5\% & 8.0\% ± 6.6\% & 21.0\% ± 18.2\% \\
tg & 63.3\% ± 3.1\% & 66.4\% ± 2.1\% & 61.3\% ± 3.8\% & 60.9\% ± 2.2\% & 3.0\% ± 2.3\% & 15.6\% ± 12.3\% \\
th & 63.7\% ± 2.6\% & 62.3\% ± 2.2\% & 62.1\% ± 3.4\% & 56.5\% ± 3.7\% & 24.6\% ± 11.3\% & 28.7\% ± 2.0\% \\
tk & 59.4\% ± 2.8\% & 57.5\% ± 3.1\% & 52.5\% ± 3.2\% & 56.1\% ± 1.8\% & 13.6\% ± 3.4\% & 7.5\% ± 10.1\% \\
tl & 73.4\% ± 3.4\% & 70.0\% ± 2.6\% & 67.1\% ± 4.7\% & 64.3\% ± 2.7\% & 28.6\% ± 2.9\% & 30.1\% ± 2.7\% \\
tn & 29.1\% ± 12.9\% & 33.8\% ± 14.9\% & 18.5\% ± 15.6\% & 43.7\% ± 5.0\% & 6.9\% ± 5.7\% & 23.6\% ± 4.5\% \\
tr & 74.1\% ± 3.0\% & 64.3\% ± 2.0\% & 69.0\% ± 5.5\% & 69.4\% ± 1.3\% & 29.3\% ± 4.0\% & 28.7\% ± 4.6\% \\
trv & 35.1\% ± 2.7\% & 37.7\% ± 2.5\% & 19.6\% ± 6.4\% & 30.8\% ± 5.2\% & 14.6\% ± 7.1\% & 22.3\% ± 4.8\% \\
ts & 29.7\% ± 9.2\% & 30.8\% ± 8.9\% & 13.7\% ± 5.9\% & 34.8\% ± 7.6\% & 10.4\% ± 10.9\% & 19.1\% ± 20.5\% \\
tt & 67.4\% ± 21.8\% & 64.1\% ± 20.5\% & 49.6\% ± 16.6\% & 60.0\% ± 3.8\% & 39.3\% ± 2.8\% & 43.7\% ± 4.6\% \\
tum & 40.4\% ± 11.5\% & 53.3\% ± 7.2\% & 18.3\% ± 8.7\% & 40.5\% ± 7.2\% & 16.5\% ± 4.0\% & 27.7\% ± 12.0\% \\
tw & 35.8\% ± 3.2\% & 39.4\% ± 2.4\% & 31.2\% ± 8.9\% & 45.0\% ± 2.9\% & 1.7\% ± 1.7\% & 14.7\% ± 12.1\% \\
tyv & 44.8\% ± 3.8\% & 47.7\% ± 2.6\% & 46.2\% ± 5.7\% & 49.9\% ± 2.5\% & 11.5\% ± 5.3\% & 19.1\% ± 9.1\% \\
udm & 57.0\% ± 6.1\% & 60.9\% ± 3.4\% & 51.4\% ± 9.0\% & 58.9\% ± 3.2\% & 31.1\% ± 3.0\% & 30.2\% ± 4.6\% \\
ug & 43.8\% ± 11.5\% & 45.3\% ± 4.4\% & 42.9\% ± 5.1\% & 46.9\% ± 3.7\% & 22.0\% ± 4.4\% & 20.1\% ± 9.2\% \\
uk & 64.6\% ± 6.0\% & 61.5\% ± 2.9\% & 61.4\% ± 7.1\% & 52.8\% ± 2.1\% & 29.9\% ± 2.4\% & 30.5\% ± 3.3\% \\
ur & 66.7\% ± 6.7\% & 65.4\% ± 4.7\% & 63.2\% ± 6.7\% & 64.9\% ± 3.5\% & 28.8\% ± 2.0\% & 27.3\% ± 3.9\% \\
uz & 61.7\% ± 6.8\% & 57.5\% ± 2.8\% & 50.9\% ± 3.1\% & 56.4\% ± 2.0\% & 27.5\% ± 2.4\% & 27.2\% ± 2.5\% \\
vec & 67.3\% ± 4.0\% & 68.4\% ± 3.0\% & 60.6\% ± 4.0\% & 56.5\% ± 4.2\% & 17.2\% ± 4.6\% & 26.6\% ± 5.7\% \\
vep & 48.6\% ± 3.1\% & 53.0\% ± 1.4\% & 44.7\% ± 3.0\% & 51.3\% ± 2.1\% & 25.6\% ± 2.9\% & 24.8\% ± 11.7\% \\
vi & 78.9\% ± 5.5\% & 75.8\% ± 4.5\% & 72.8\% ± 7.3\% & 67.7\% ± 3.7\% & 30.1\% ± 3.2\% & 30.1\% ± 2.7\% \\
vls & 70.6\% ± 4.5\% & 69.3\% ± 3.5\% & 63.8\% ± 4.6\% & 58.0\% ± 2.9\% & 21.4\% ± 3.1\% & 26.1\% ± 11.4\% \\
vo & 60.2\% ± 5.4\% & 58.5\% ± 5.6\% & 13.4\% ± 10.9\% & 52.1\% ± 7.5\% & 40.3\% ± 3.9\% & 42.1\% ± 2.6\% \\
wa & 46.9\% ± 3.6\% & 48.1\% ± 2.4\% & 43.4\% ± 4.2\% & 48.7\% ± 2.3\% & 5.6\% ± 4.4\% & 6.1\% ± 9.6\% \\
war & 48.1\% ± 16.6\% & 58.5\% ± 8.9\% & 32.6\% ± 11.7\% & 48.8\% ± 5.3\% & 53.1\% ± 2.7\% & 59.0\% ± 4.1\% \\
wo & 22.6\% ± 6.8\% & 25.3\% ± 5.0\% & 15.7\% ± 4.8\% & 24.3\% ± 3.6\% & 2.5\% ± 3.8\% & 6.4\% ± 10.2\% \\
wuu & 43.8\% ± 6.9\% & 42.6\% ± 4.7\% & 40.2\% ± 4.0\% & 41.1\% ± 1.8\% & 16.5\% ± 2.9\% & 15.6\% ± 7.8\% \\
xh & 37.0\% ± 6.3\% & 43.4\% ± 5.3\% & 18.3\% ± 6.7\% & 24.1\% ± 3.9\% & 5.1\% ± 3.2\% & 13.5\% ± 6.2\% \\
xmf & 49.7\% ± 2.6\% & 50.0\% ± 3.0\% & 43.9\% ± 2.0\% & 45.1\% ± 1.1\% & 20.1\% ± 3.1\% & 21.5\% ± 9.8\% \\
yi & 59.2\% ± 6.8\% & 62.2\% ± 2.7\% & 55.8\% ± 5.1\% & 57.1\% ± 2.9\% & 26.1\% ± 3.4\% & 29.7\% ± 4.3\% \\
yo & 39.4\% ± 5.0\% & 45.2\% ± 7.3\% & 42.2\% ± 4.8\% & 51.0\% ± 3.3\% & 0.1\% ± 0.3\% & 16.0\% ± 8.6\% \\
yue & 56.3\% ± 3.5\% & 51.2\% ± 2.6\% & 48.1\% ± 5.1\% & 52.7\% ± 2.3\% & 12.2\% ± 8.4\% & 15.6\% ± 6.9\% \\
zea & 71.8\% ± 2.5\% & 69.5\% ± 2.9\% & 62.8\% ± 5.6\% & 67.0\% ± 2.3\% & 23.3\% ± 3.4\% & 29.1\% ± 3.6\% \\
zh-cn & 54.3\% ± 2.9\% & 47.0\% ± 2.8\% & 51.1\% ± 4.9\% & 49.2\% ± 3.6\% & 8.8\% ± 11.3\% & 16.1\% ± 8.7\% \\
zh-tw & 58.1\% ± 4.3\% & 48.7\% ± 2.7\% & 52.4\% ± 3.0\% & 54.9\% ± 2.4\% & 15.3\% ± 7.1\% & 14.3\% ± 11.1\% \\
zu & 37.9\% ± 5.0\% & 46.8\% ± 4.8\% & 27.2\% ± 4.8\% & 34.0\% ± 2.4\% & 11.8\% ± 6.6\% & 14.2\% ± 11.3\% \\
\hline
\end{tabular}
\caption{The F1-scores on MultiWikiQA for languages 196-261, sorted alphabetically.}
\label{tab:num-samples-in-small-subsets}
\end{table*}

\begin{figure*}
    \centering
    \includegraphics[width=1.0\linewidth]{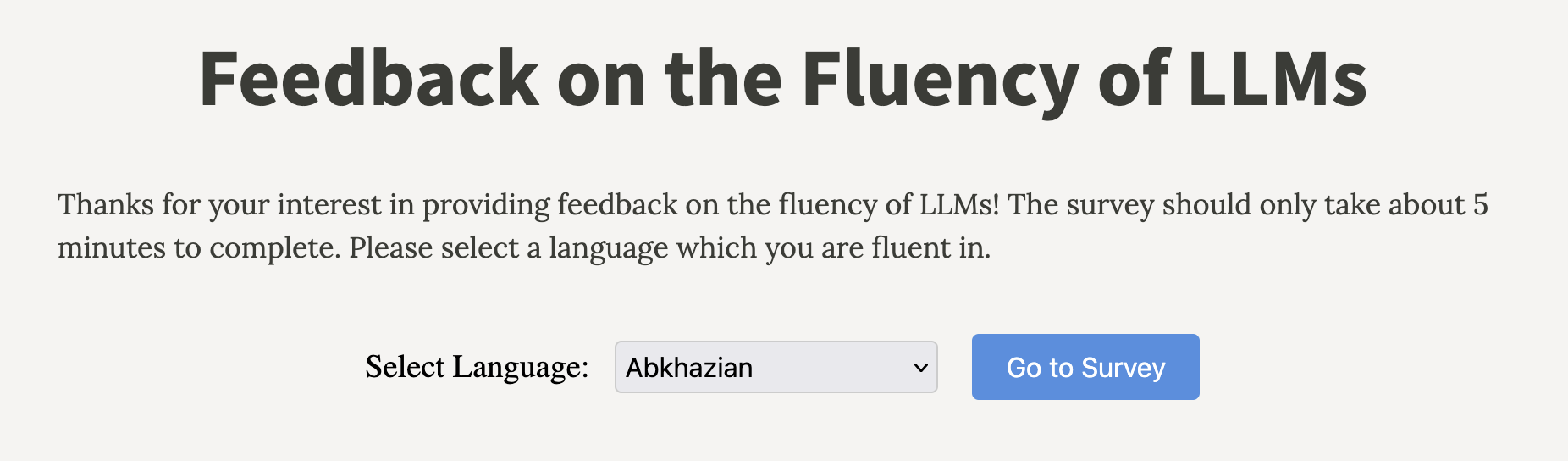}
    \caption{The survey routing interface.}
    \label{fig:routing-interface}
\end{figure*}

\begin{figure*}[h]
\scriptsize
\centering
\begin{tabular}{c}
\begin{lstlisting}
<script lang="ts" setup>
const languageToSurveyUrl: Record<string, string> = {
  Abkhazian: "https://forms.cloud.microsoft/e/pdmAKbsRk1",
  Acehnese: "",
  (...)
};

function goToSurvey() {
  const selectElement = document.getElementById(
    "language-select",
  ) as HTMLSelectElement;
  const selectedLanguage = selectElement.value;
  const surveyUrl = languageToSurveyUrl[selectedLanguage];
  if (surveyUrl) {
    window.open(surveyUrl, "_blank");
  } else {
    window.open(
      "mailto:[redacted]@[redacted].[redacted]?" +
        "subject=Fluency%20Survey%20Language%20Support - " +
        selectedLanguage +
        "&" +
        "body=I would like to request support for " +
        selectedLanguage +
        " in the fluency survey. Thanks!",
      "_blank",
    );
  }
}
</script>
<template>
  <h1 class="centered">Feedback on the Fluency of LLMs</h1>
  <p class="centered-box serif-text">
    Thanks for your interest in providing feedback on the fluency of LLMs! The
    survey should only take about 5 minutes to complete. Please select a
    language which you are fluent in.
  </p>

  <br />
  <br />

  <div class="centered">
    <label for="language-select" class="language-label"
      >Select Language:
    </label>
    <select id="language-select" class="dropdown">
      <option
        v-for="language in Object.keys(languageToSurveyUrl)"
        v-bind:key="language"
        :value="language"
      >
        {{ language }}
      </option>
    </select>
    <button class="button" @click="goToSurvey">Go to Survey</button>
  </div>
</template>
<style scoped>
h1 {
  font-size: 3rem;
}
.centered {
  text-align: center;
}
.language-label {
  font-size: 1.2rem;
  margin-right: 10px;
}
.dropdown {
  font-size: 1rem;
  padding: 5px;
  margin-right: 20px;
  border: 1px solid #ccc;
  border-radius: 4px;
}
.button {
  font-size: 1rem;
  padding: 10px 20px;
  color: white;
  background-color: #4a90e2;
  border: none;
  border-radius: 4px;
  cursor: pointer;
}
.button:hover {
  background-color: #357abd;
}
</style>
\end{lstlisting}
\end{tabular}
\caption{The source code for the Vue.js survey routing interface component}
\end{figure*}

\end{document}